\documentclass[letterpaper]{article} 
\usepackage{aaai2026}
\usepackage{times}  
\usepackage{helvet}  
\usepackage{courier}  
\usepackage[hyphens]{url}  
\usepackage{graphicx} 
\usepackage{multirow}   
\usepackage{makecell}
\usepackage{natbib}  
\usepackage{caption} 
\usepackage{algorithm}
\usepackage{algorithmicx}
\usepackage{algpseudocode}
\usepackage{amssymb}
\usepackage{amsmath}    
\usepackage{bm}         
\usepackage{tabularx}
\usepackage{listings}
\usepackage{xcolor}
\usepackage{booktabs}
\usepackage{newfloat}
\usepackage{listings}
\DeclareCaptionStyle{ruled}{labelfont=normalfont,labelsep=colon,strut=off} 
\floatstyle{ruled}
\newfloat{listing}{tb}{lst}{}
\floatname{listing}{Listing}
\newcommand{\dist}{\text{dist}}
\title{CAST-LUT: Tokenizer-Guided HSV Look-Up Tables for Purple Flare Removal}

\author {
    Pu Wang \textsuperscript{\rm 1},
    Shuning Sun\textsuperscript{\rm 2},
    Jialang Lu\textsuperscript{\rm 3},
    Chen Wu\textsuperscript{\rm 4},
    Zhihua Zhang\textsuperscript{\rm 1},
    Youshan Zhang\textsuperscript{\rm 5},
    Chenggang Shan\textsuperscript{\rm 6},
    Dianjie Lu\textsuperscript{\rm 7},
    Guijuan Zhang\textsuperscript{\rm 7},
    Zhuoran Zheng\textsuperscript{\rm 8 } 
}
\affiliations {
    \textsuperscript{\rm 1} Shandong University\\
    \textsuperscript{\rm 2}University of the Chinese Academy of Sciences\\
    \textsuperscript{\rm 3}Hubei University\\
    \textsuperscript{\rm 4}University of Science and Technology of China\\
    \textsuperscript{\rm 5} Yeshiva University\\
    \textsuperscript{\rm 6} Zaozhuang University\\
    \textsuperscript{\rm 7} Shandong Normal University\\
    \textsuperscript{\rm 8} Qilu University of Technology\\
    wangou@mail.sdu.edu.cn, sunshuning23@mails.ucas.ac.cn, lujialang@stu.hubu.edu.cn,
    wuchen5X@mail.ustc.edu.cn,
    zhangzhihua@sdu.edu.cn,
    yz945@cornell.edu,
    shanchenggang@uzz.edu.cn,
    \{ludianjie, zhangguijuan\}@sdnu.edu.cn,
    zhengzr@njust.edu.cn
    
}
\usepackage{bibentry}

\begin{document}

\maketitle
\begin{abstract}

Purple flare, a diffuse chromatic aberration artifact commonly found around highlight areas, severely degrades the tone transition and color of the image. Existing traditional methods are based on hand-crafted features, which lack flexibility and rely entirely on fixed priors,
while the scarcity of paired training data critically hampers deep learning.  To address this issue, we propose a novel network built upon decoupled HSV Look-Up Tables (LUTs). The method aims to simplify color correction by adjusting the Hue (H), Saturation (S), and Value (V) components independently. This approach resolves the inherent color coupling problems in traditional methods. Our model adopts a two-stage architecture: First, a Chroma-Aware Spectral Tokenizer (CAST) converts the input image from RGB space to HSV space and independently encodes the Hue (H) and Value (V) channels into a set of semantic tokens describing the Purple flare status; second, the HSV-LUT module takes these tokens as input and dynamically generates independent correction curves (1D-LUTs) for the three channels H, S, and V. To effectively train and validate our model, we built the first large-scale purple flare dataset with diverse scenes. We also proposed new metrics and a loss function specifically designed for this task. Extensive experiments demonstrate that our model not only significantly outperforms existing methods in visual effects but also achieves state-of-the-art performance on all quantitative metrics.
\end{abstract}

\begin{links}
    \link{Code}{https://github.com/Pu-Wang-alt/Reduce-Purple-Flare/}
    \link{Datasets}{https://huggingface.co/datasets/PuWang0/purple_flare}
    \link{Formal version}{}
\end{links}

\section{Introduction}
In complex lighting environments, images are often affected by various color artifacts~\cite{yu2017low, abbasi2024enhancing}, with the purple flare being one of the most common. Unlike general image degradations such as noise or blur, purple flare presents a unique challenge due to its diffuse nature, color specificity, and strong spatial correlation with highlight regions, severely impacting the image's color fidelity~\cite{min2025exploring, chen2024topiq}. This issue is especially prominent in mobile photography, consumer-grade cameras, and low-cost optical systems, primarily caused by factors such as internal lens reflections, sensor saturation, and chromatic dispersion~\cite{spencer1995physically}.

\begin{figure}[t]
\centering
\includegraphics[width=0.9\linewidth]{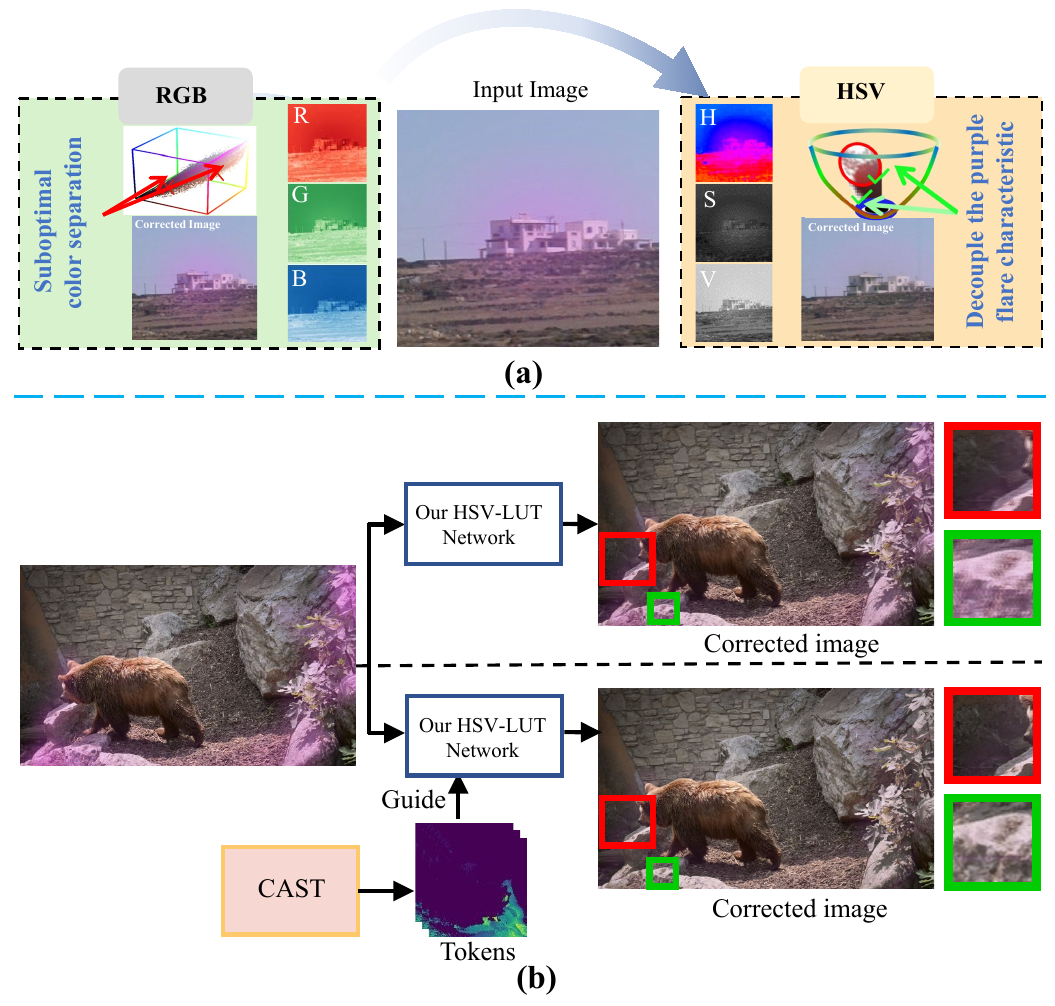} 
\caption{
\textbf{Color space analysis and CAST-guided correction comparison.} (a) RGB vs. HSV: HSV decouples purple flare characteristics more clearly. 
(b) CAST-guided effect: The CAST module significantly improves performance.
}
\label{fig1}
\vspace{-4mm}
\end{figure}

Recently, image enhancement techniques have made significant progress in areas such as deraining~\cite{liu2025deraings}, dehazing~\cite{sabitha2024restoration, kumari2024real, agrawal2024fog}, and color restoration~\cite{chen2021pre, liang2021swinir}. However, the purple flare has not been systematically studied yet. Early attempts relied on traditional methods utilizing decoupled color spaces like HSV or Lab~\cite{lee2011purple,kim2010detection,ware2018measuring}. These methods are constrained by their reliance on hand-crafted, fixed thresholds, which causes them to lack flexibility and generalization capability in complex scenes, limiting their practical application.
To overcome the limitations of traditional methods, researchers have turned to deep learning. Wu et al.~\shortcite{wu2021train} pioneered the application of end-to-end CNNs to general flare removal by constructing a large-scale semi-synthetic dataset.  
To further bridge the domain gap between synthetic data and real-world flares, Deng et al.~\shortcite{deng2024towards} proposed a knowledge-driven hybrid approach that utilizes a flare-level estimator as prior knowledge to guide the restoration network.  
While these methods represent the state-of-the-art, they are essentially pixel-to-pixel restoration networks operating in the coupled RGB color space, lacking precise control over color decoupling for specific chromatic artifacts. Among them, driven by the need for efficiency, LUT-based methods have emerged as a prominent choice for lightweight architectures. However, existing LUT methods rely on the RGB space and are not sensitive to capturing the purple flare of local region transformation (see Fig.~\ref{fig1}(a)). We observed that the HSV space can amplify the characteristic of local purple flare. Still, the non-linear transformation from RGB to HSV can amplify local color space noise, creating new artifacts in the final result~\cite{gevers2012color}. 
To leverage the benefits of the decoupled HSV color space without incurring the instability of traditional conversion, we propose CAST-LUT. This method employs a Chroma-Aware Spectral Tokenizer (CAST) to generate stable, high-level semantic representations of the purple flare.
Fig.~\ref{fig1}(b) illustrates the critical guiding role of the CAST module. These semantic tokens then guide the generation of independent 1D-LUTs for the H, S, and V channels. 

Additionally, we constructed the Purple Flare Synthesis Dataset (PFSD), comprising 4,987 training pairs, 608 validation pairs, and 618 testing pairs.  To simulate optical lens characteristics, we identify highlight regions at object contour edges as candidates, apply a Gaussian blur, and fuse purple flare into raw images.  This is the first large-scale paired image collection for this task.  Due to the localized nature of purple flare, averaging errors across the entire image can mask residual artifacts in small regions, making traditional global metrics ineffective for accurate evaluation.  We therefore introduce new metrics: using a purple flare mask, we separate flare-region PSNR (PSNR-F) and non-flare-region PSNR (PSNR-NF) to assess flare restoration and clean-region fidelity, respectively.  For color correction, we propose Hue Alignment Error (HAE), which quantifies hue differences in flare regions to evaluate color recovery.

\begin{itemize}
\item 
We propose CAST-LUT, which avoids traditional RGB-to-HSV noise amplification. Its Chroma-Aware Spectral Tokenizer (CAST) diagnoses purple flare into semantic tokens that guide an adaptive 1D-LUT for precise correction of each HSV channel.
\item To address the data scarcity for the purple flare removal task, we construct a large-scale, high-quality purple flare dataset containing diverse scenes and devices, providing a fair benchmark for training and evaluation.
\item 
We introduce a new loss function and metrics for accurate evaluation. PSNR-F/NF assesses flare removal and detail preservation, respectively, while our proposed Hue Alignment Error (HAE) metric measures color fidelity.

\end{itemize}

\section{Related Work}
\textbf{Chromatic Aberration and Flare Removal.}
Traditional methods note limitations in processing purple flare directly in sRGB space, as intense luminance variations interfere with color assessment. Researchers have explored other color spaces~\cite{malik2018iterative}. Chung et al.~\shortcite{chung2009detecting} identified chromatic fringes by comparing R, G, B channel intensities but faced false positives due to ambiguous purple definitions. To better separate chrominance from luminance, subsequent works shifted to other color spaces. Kim et al.~\shortcite{kim2010detection} used CIExy for luminance normalization, while Ju et al. leveraged YCbCr's chrominance channels (Cb-Cr) to analyze color deviations independently~\cite{ju2013colour}. However, a common bottleneck is their reliance on static, experimentally determined absolute thresholds to filter high-contrast or near-saturated areas. Given varied purple flare characteristics across cameras and scenes, fixed thresholds lack universality and robustness. Considering degraded image representation reconstruction ability, we propose using a word segmenter to aid in detail reconstruction.

\noindent \textbf{Discrete Semantic Representation with Vector Quantization.}
Vector Quantization (VQ) is widely used to generate semantic tokens, a key component of modern perceptual systems.
Its core principle maps high-dimensional feature vectors to a finite set of learned codebook embeddings. This technique gained prominence with VQ-VAE~\cite{van2017neural}, which established a discrete latent space in an autoencoder architecture and integrated a discrete latent bottleneck into such frameworks. A discrete latent space offers significant advantages: it enables learning rich, semantically meaningful representations, where each codebook entry captures distinct perceptual patterns analogous to vocabulary words~\cite{schwettmann2021toward}. Additionally, discretization improves generative model stability and efficiency by mitigating issues like posterior collapse, a challenge in traditional VAEs~\cite{xue2019supervised}. The discretization of visual data into tokens has enabled powerful architectures like Transformers~\cite{han2022survey} and inspired token or prompt-guided approaches for other vision tasks, such as few-shot learning \cite{li2024knn, li2025vt}.
\begin{figure*}[t]
\centering
\includegraphics[width=0.95\textwidth]{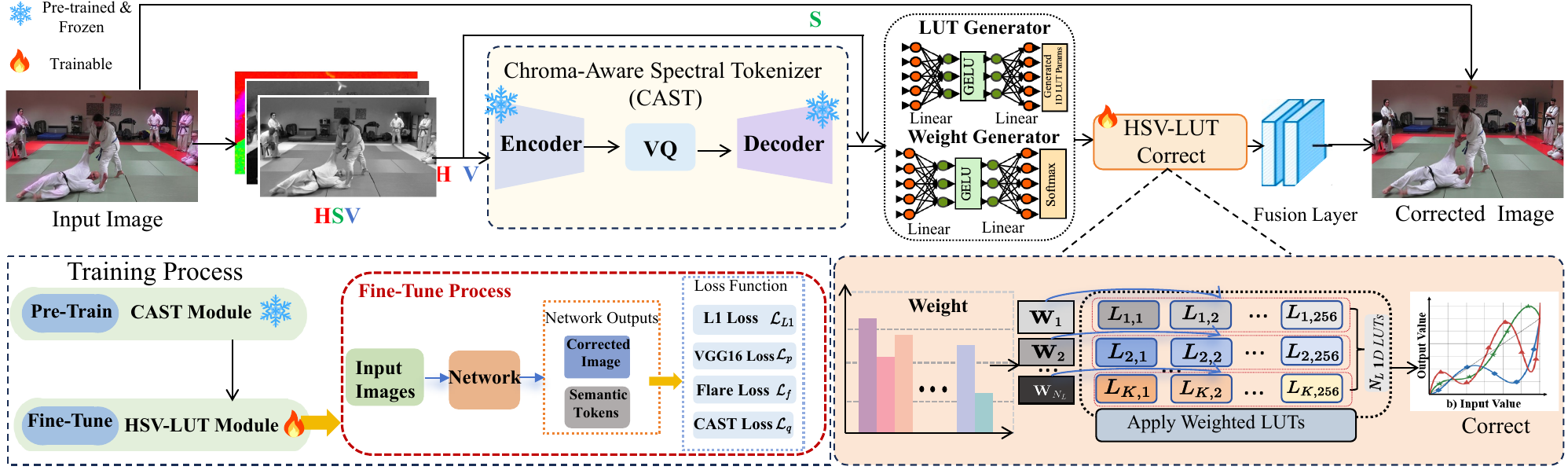} 
\caption{Overall architecture of the decoupled CAST-LUT network. The CAST module first extracts semantic tokens from the input image's H and V channels via an Encoder-VQ pipeline. These tokens then guide the dynamic generation and weighted fusion of multiple, decoupled 1D-LUTs for the H, S, and V channels to obtain the final corrected image.}
\label{fig2}
\end{figure*}

\noindent \textbf{LUT-based Color Image Enhancement.} 
Recently, the LUT paradigm has shown significant potential across image restoration tasks, evolving into complex task-specific architectures. For color image denoising, where high color fidelity is critical, early LUT methods were limited by independent color channel processing. Therefore, DnLUT~\cite{yang2025dnlut} introduced a Pairwise Channel Mixer, which explicitly models inter-channel correlations by processing RGB channel pairs in parallel, boosting color denoising performance. For complex mixed-degradation scenarios (e.g., low-light enhancement), the LUT framework has advanced further. DPLUT~\cite{lin2025dplut} proposed a two-stage approach, using a Light-Adjustment LUT and Noise-Suppression LUT to decouple brightening and denoising sub-tasks. However, despite SOTA performance in their domains, these advanced LUT methods are not designed for purple flare removal. 
Their model priors are inconsistent with the unique characteristics and spatial distribution of flare, and they are confined to operating in the RGB space.

\section{Methodology}
Our proposed purple flare removal network, named \textbf{Decoupled HSV-LUT Network}, adopts a novel two-stage ``perceive-then-correct" paradigm. 
An overview of our framework is illustrated in Figure \ref{fig2}.

\subsection{Chroma-Aware Spectral Tokenizer (CAST)}
The CAST module acts as the framework’s perception engine, tasked with diagnosing purple flare features by translating complex visual artifacts into discrete semantic tokens.

Recognizing that purple flare is characterized by abnormal hue in highlight regions, we adopt a targeted approach. The $\textbf{H}$ channel (chromatic signature) and $\textbf{V}$ channel (highlight spatial context) undergo 4$\times$ downsampling via a shared-weight CNN encoder, which processes them independently to extract hierarchical feature maps $\mathbf{F}_\textbf{H}, \mathbf{F}_\textbf{V} \in \mathbb{R}^{8 \times 256 \times H' \times W'}$. The $\textbf{S}$ channel undergoes no encoding processing and remains in its original state to preserve the image’s inherent color richness.  The encoder extracts hierarchical feature maps for each channel:
\begin{equation}
    \mathbf{F}_\textbf{H} = \text{Encoder}(\mathbf{H}), \quad \mathbf{F}_\textbf{V} = \text{Encoder}(\mathbf{V}).
\end{equation}
To bridge low-level visual features and high-level semantic understanding, these continuous feature maps $\mathbf{F}_\textbf{H}$ and $\mathbf{F}_\textbf{V}$ are discretized into semantic tokens via a Vector Quantization (VQ) module, where specific tokens learn to represent recurring purple flare characteristics (e.g., ``high-intensity purple on edges"). Specifically, the VQ module maps each feature vector to the nearest codebook entry, thereby effectively categorizing visual patterns and generating discrete token indices and quantized representations:
\begin{equation}
    \mathbf{T}_\textbf{H}, \mathbf{F}_{\text{quant}}^{\textbf{H}} = \text{VQ}(\mathbf{F}_\textbf{H}), \quad \mathbf{T}_\textbf{V}, \mathbf{F}_{\text{quant}}^{\textbf{V}} = \text{VQ}(\mathbf{F}_\textbf{V}).
\end{equation}
Next, a decoder network reconstructs the H and V channels from the quantized features.
\begin{equation}
    \hat{\mathbf{H}} = \text{Decoder}(\mathbf{F}_{\text{quant}}^{\textbf{H}}), \quad \hat{\mathbf{V}} = \text{Decoder}(\mathbf{F}_{\text{quant}}^{\textbf{V}}).
    \label{eq:decoder}
\end{equation}
The original \textbf{S} channel is combined with $\hat{\mathbf{H}}$ and $\hat{\mathbf{V}}$ to form a reconstructed HSV image, which is then converted back to RGB to obtain the initial reconstructed image $\mathbf{I}_{\text{recon}}^{\text{RGB}}$. The training of CAST is supervised by the reconstruction error between $\mathbf{I}_{\text{recon}}^{\text{RGB}}$ and $\mathbf{I}^{\text{RGB}}$, ensuring that quantized features preserve key information from the input image. In the subsequent color correction stage, we treat the discrete token indices $\mathbf{T}_{\text{combined}}=[\mathbf{T}_H, \mathbf{T}_V]$ as high-level semantic representations of purple flare characteristics and feed them into the LUT module.  


\subsection{HSV-LUT Color Correction}
The color correction step of our method is executed by the HSV-based decoupled 1D-LUT module. Guided by the semantic tokens $\mathbf{T}_{\text{combined}}$ generated by CAST, this module performs color correction on the reconstructed image $\mathbf{I}_{\text{recon}}^{\text{RGB}}$ while leveraging residual information from the original image $\mathbf{I}^{\text{RGB}}$ to preserve high-frequency details.

\noindent \textbf{Token-Guided LUT Generation.}
First, the semantic tokens $\mathbf{T}_{\text{combined}}$ are converted into a global feature vector $\mathbf{f}_{\text{token}}$ containing purple flare characteristics through embedding and aggregation operations, which drives two parallel sub-networks: the LUT generator and the weight generator.
\begin{equation}
    \mathbf{f}_{\text{token}} = \text{Aggregate}(\text{Embed}(\mathbf{T}_{\text{combined}})).
\end{equation}
(1) \textbf{LUT Generator:} This network processes $\mathbf{f}_{\text{token}}$ using an MLP to generate and reshape parameters $\mathbf{P}$ for $N_L$ groups of LUTs. Each group contains three independent 1D-LUTs tailored to the H, S, and V channels, respectively. This decoupled design enables targeted adjustments for each color attribute.
\begin{equation}
        \mathbf{P} = \text{Reshape}(\text{MLP}(\mathbf{f}_{\text{token}})).
\end{equation}
(2) \textbf{Weight Generator.} Parallel to the LUT generator, the weight generator processes $\mathbf{f}_{\text{token}}$ using an MLP and normalizes via Softmax to predict dynamic fusion weights $W \in \mathbb{R}^{N_L}$. These weights determine the contribution of the $N_L$ LUT groups.  
\begin{equation}
\mathbf{W} = \text{Softmax}(\text{MLP}(\mathbf{f}_{\text{token}})).
\end{equation}

\noindent\textbf{Decoupled HSV Correction via 1D-LUTs.}
First, the reconstructed image $\mathbf{I}_{\text{recon}}^{\text{RGB}}$ generated by CAST is converted to the HSV color space and decomposed into its three channels $[\mathbf{H}_\text{in}, \mathbf{S}_\text{in}, \mathbf{V}_\text{in}]$.

Our correction process operates on each channel independently within the HSV color space. For each LUT set $i=1, \dots, N_L$, its corresponding three 1D-LUTs ($\text{LUT}_{\textbf{H},i}, \text{LUT}_{\textbf{S},i}, \text{LUT}_{\textbf{V},i}$) are applied to the three input channels $[\mathbf{H}_\text{in}, \mathbf{S}_\text{in}, \mathbf{V}_\text{in}]$ respectively. These results are then fused into the final corrected channels $\mathbf{C}_\text{final}$ via weighted averaging with the predicted weight vector $\mathbf{W}$. This process can be described by the following equation:  
\begin{equation}
    \mathbf{C}_\text{final} = \sum_{i=1}^{N_L} \mathbf{W}_i \cdot \text{ApplyLUT}(\mathbf{C}_\text{in}, \text{LUT}_{\textbf{C},i}), 
\end{equation}
where $\mathbf{C}_\text{in} \in \{\mathbf{H}_\text{in}, \mathbf{S}_\text{in}, \mathbf{V}_\text{in}\}$.
Finally,  the final corrected channels $\mathbf{C}_\text{final}$ are re-combined and converted back to the RGB color space, yielding the LUT-corrected image $\mathbf{I}_{\text{fused}}^{\text{RGB}}$. 

\noindent \textbf{Residual Branch and Final Fusion.}
To preserve details, a residual branch processes the original input image $\mathbf{I}^{\text{RGB}}$ to extract complementary high-frequency features, denoted as $\mathbf{I}_{\text{residual}}$. The color-corrected image from the main path, $\mathbf{I}_{\text{fused}}^{\text{RGB}}$, is then concatenated with these residual features to seamlessly integrate the information from both paths. A global skip connection adds the original image $\mathbf{I}^{\text{RGB}}$ to this fused result, producing the final corrected image $\mathbf{I}_{\text{output}}$.
\begin{equation}
\mathbf{I}_{\text{output}} = \text{Fusion}(\text{Concat}(\mathbf{I}_{\text{fused}}^{\text{RGB}}, \mathbf{I}_{\text{residual}})) + \mathbf{I}^{\text{RGB}}.
\end{equation}
This architecture stabilizes training and focuses the network on learning necessary corrections to the residual instead of reconstructing the entire image from scratch. This approach stabilizes training and allows the two branches to specialize: the main branch focuses on complex color correction, while the residual branch ensures high-fidelity detail preservation.

\subsection{Loss Function}
Our model is trained using a composite loss function $\mathcal{L}_{total}$:
\begin{equation}
\mathcal{L}_{total} = \lambda_{1} \mathcal{L}_{L1} + \lambda_{p} \mathcal{L}_{p} + \lambda_{f} \mathcal{L}_{f} + \lambda_{q} \mathcal{L}_{q},
\end{equation}
where $\lambda_{1}, \lambda_{p}, \lambda_{f}, \lambda_{q}$ balance contributions from pixel accuracy, perceptual similarity, artifact suppression, and representation stability. $\mathcal{L}_{L1} = ||\mathbf{I}_\text{output} - \mathbf{I}_\text{GT} ||_1$ ensures pixel-level fidelity, while $\mathcal{L}_{p} = || \phi(\mathbf{I}_\text{output}) - \phi(\mathbf{I}_\text{GT})||_1$ uses VGG-16 features~\cite{simonyan2014very} to enhance perceptual quality.   Finally, $\mathcal{L}_{q} = ||\text{sg}(\mathbf{F}) - \mathbf{e}||_2^2$ regularizes the codebook by penalizing the distance between the encoder's output feature $\mathbf{F}$, and its nearest codebook vector $\mathbf{e}$, ensuring stable token representations.

\begin{figure}[t]
    \centering
    \includegraphics[width=0.95\columnwidth]{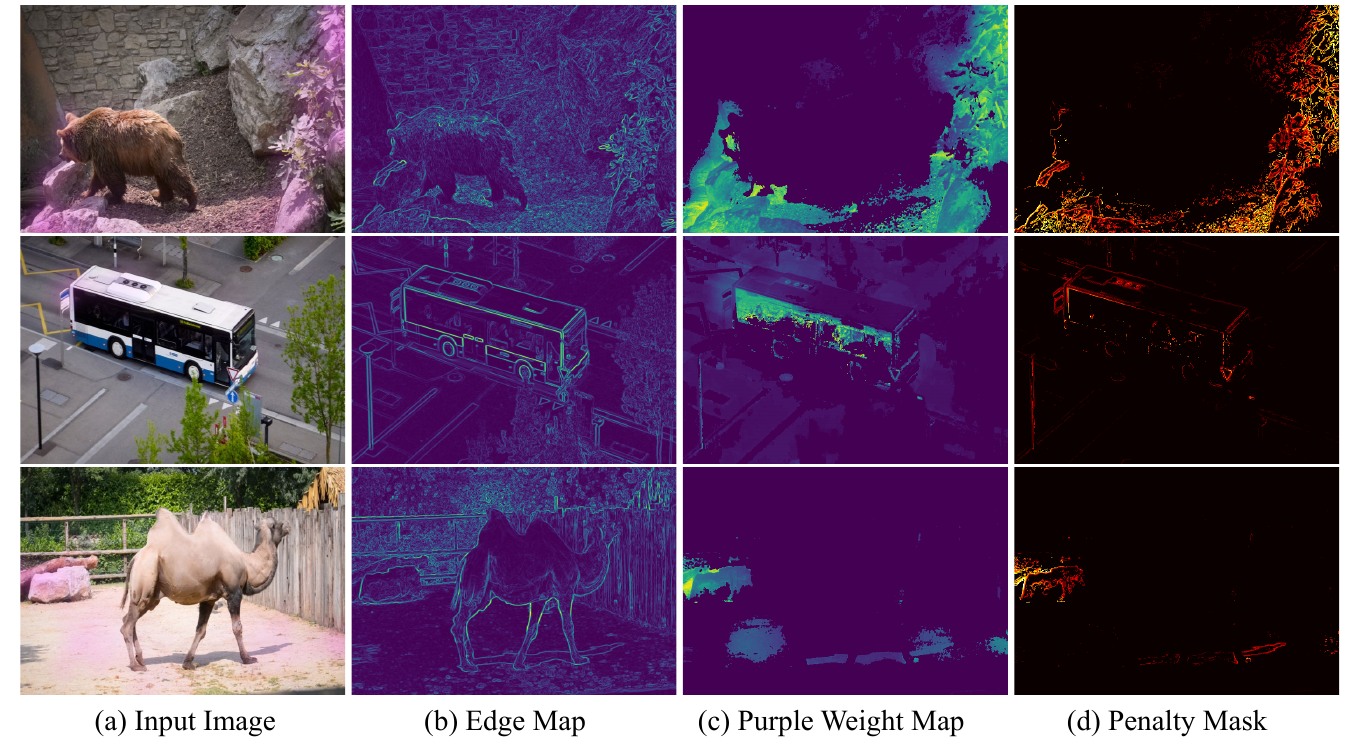} 
    \caption{The generation process of the penalty mask $\mathbf{M}$ used in our flare suppression loss $\mathcal{L}_f$.   The final mask (d) is obtained by element-wise multiplication of the edge map (b) and the purple weight map (c), which can precisely locate flare artifacts and apply targeted loss.}
    \label{fig:mask}
    \vspace{-4mm}
\end{figure}

\noindent \textbf{Purple Flare Suppression Loss $\mathcal{L}_{f}$.}
We have introduced a purple flare suppression loss specifically targeting the purple flare.
Standard L1 or L2 losses treat errors across all pixels equally, failing to focus on addressing specific color artifacts in targeted regions. To tackle this, we introduce a weighted loss term that amplifies errors in purple flare regions through a penalty mask $\mathbf{M}$, as shown in Figure \ref{fig:mask}. This mask is obtained by element-wise multiplication of two components:  
\begin{equation}
    \mathbf{M} = \mathbf{M}_\text{flare} \odot \mathbf{M}_\text{edge},
\end{equation}
where $\odot$ denotes element-wise multiplication, $\mathbf{M}_\text{flare}$ is a color feature map used to identify pixels in the image with purple hue and high saturation, $\mathbf{M}_\text{edge}$ is an edge map generated by edge detection operators such as Sobel. In this way, the loss is significantly amplified only in regions that simultaneously satisfy the two conditions of ``purple color" and ``being located at edges". The final suppression loss is defined as:  
\begin{equation}
    \mathcal{L}_f = || \mathbf{M} \odot (\mathbf{I}_\text{output} - \mathbf{I}_\text{GT}) ||_1.
\end{equation}

\begin{table*}[t]
 \centering
 \footnotesize
 \setlength{\tabcolsep}{2pt} 
 \resizebox{\textwidth}{!}{
 \begin{tabular}{l l l cccc ccc cc}
  \toprule
  \multirow{2}{*}{Category} & \multirow{2}{*}{Method} & \multirow{2}{*}{\makecell{Venue \& \\ Year}} & \multicolumn{4}{c}{\makecell{Global Quality \\ Metrics}} & \multicolumn{3}{c}{\makecell{Artifact Specific \\ Metrics}} & \multicolumn{2}{c}{\makecell{Efficiency \\ Metrics \\($256^2$)}} \\
  \cmidrule(lr){4-7} \cmidrule(lr){8-10} \cmidrule(lr){11-12}
  & & & PSNR $\uparrow$ & SSIM $\uparrow$ & LPIPS $\downarrow$ & $\Delta E$ $\downarrow$ & PSNR-F $\uparrow$ & PSNR-NF $\uparrow$ & HAE $\downarrow$ & \makecell{FLOPs \\ (G) $\downarrow$} & \makecell{Runtime \\ (ms) $\downarrow$} \\
  \midrule
  \multirow{5}{*}{LUT-based} & 3DLUT~\cite{zeng2020learning} & TPAMI'20 & 30.34 & 0.96 & 0.06 & 4.17 & 23.87 & 30.34 & 19.12  & 26.34 & 9.36 \\
  & SR-LUT~\cite{jo2021practical} & CVPR'21 & 29.76 & 0.85 & 0.12 & 5.66 & 21.35 & 29.58 & 10.21  & 25.39 & 8.25 \\
  & SPF-LUT~\cite{li2024look} & CVPR'24 & 32.19 & 0.94 & 0.09 & 5.18 & 21.93 & 33.19 & 7.57  & 34.38 & 9.12 \\
  & NILUT~\cite{conde2024nilut} & AAAI'24 & 32.31 & 0.95 & 0.07 & 4.50 & 24.86 & 32.93 & 4.87  & 30.89 & 7.12 \\
    & DnLUT~\cite{yang2025dnlut} & CVPR'25 & 30.52 & 0.88 & 0.10 & 5.17 & 23.31 & 30.25 & 9.51  & 25.16 & 8.71 \\
  & CAST-LUT (Ours) & - & \textbf{34.96} & \textbf{0.99} & \textbf{0.03} & \textbf{2.71} & \textbf{30.74} & \textbf{34.35} & \textbf{4.10}  & \textbf{23.32} & \textbf{6.09} \\
  \midrule
  \multirow{2}{*}{Classical} & CBM3D~\cite{dabov2007color} & TIP'07 & 30.57 & 0.90 & 0.22 & 4.73 & 27.49 & 31.57 & 18.53  & - & 43.11 \\
  & MC-WNNM~\cite{xu2017multi} & ICCV'17 & 27.98 & 0.80 & 0.20 & 9.53 & 26.23 & 28.00 & 18.01  & - & 121.23 \\
  \midrule
  \multirow{2}{*}{\makecell{Scene\\Specific \\ Restoration}} 
  & Zero-DCE~\cite{guo2020zero} & CVPR'20 & 14.92 & 0.84 & 0.13 & 19.59 & 13.34 & 14.92 & 40.16  & 34.79 & 10.73 \\
  & RUAS~\cite{liu2021retinex} & CVPR'21 & 19.96 & 0.68 & 0.35 & 31.30 & 8.73 & 9.96 & 28.49  & 43.84 & 11.34 \\
  & 
LightenDiffusion~\cite{Jiang_2024_ECCV} & ECCV'24 & 30.59 & 0.97 & 0.08 & 5.01 & 24.66 & 30.59 & 27.25  & 27.01 & 10.25 \\
  & BPAM~\cite{lou2025learning} & ICCV'25 & 29.93 & 0.95 & 0.12 & 5.10 & 27.26 & 28.93 & 24.75  & 15.21& 13.04 \\
  & HVI-CIDNet~\cite{yan2025hvi} & CVPR'25 & 32.17 & 0.98 & 0.04 & 2.92 & 26.10 & 33.17 & 17.39  & 27.50 & 14.94 \\
  \bottomrule
 \end{tabular}}
 \caption{Quantitative comparison with state-of-the-art methods on our purple flare dataset. $\uparrow$ indicates higher is better, and $\downarrow$ indicates lower is better.
 }
\label{tab:sota_comparison}
\end{table*}

\begin{figure*}[h!]
    \centering
    \includegraphics[width=\textwidth]{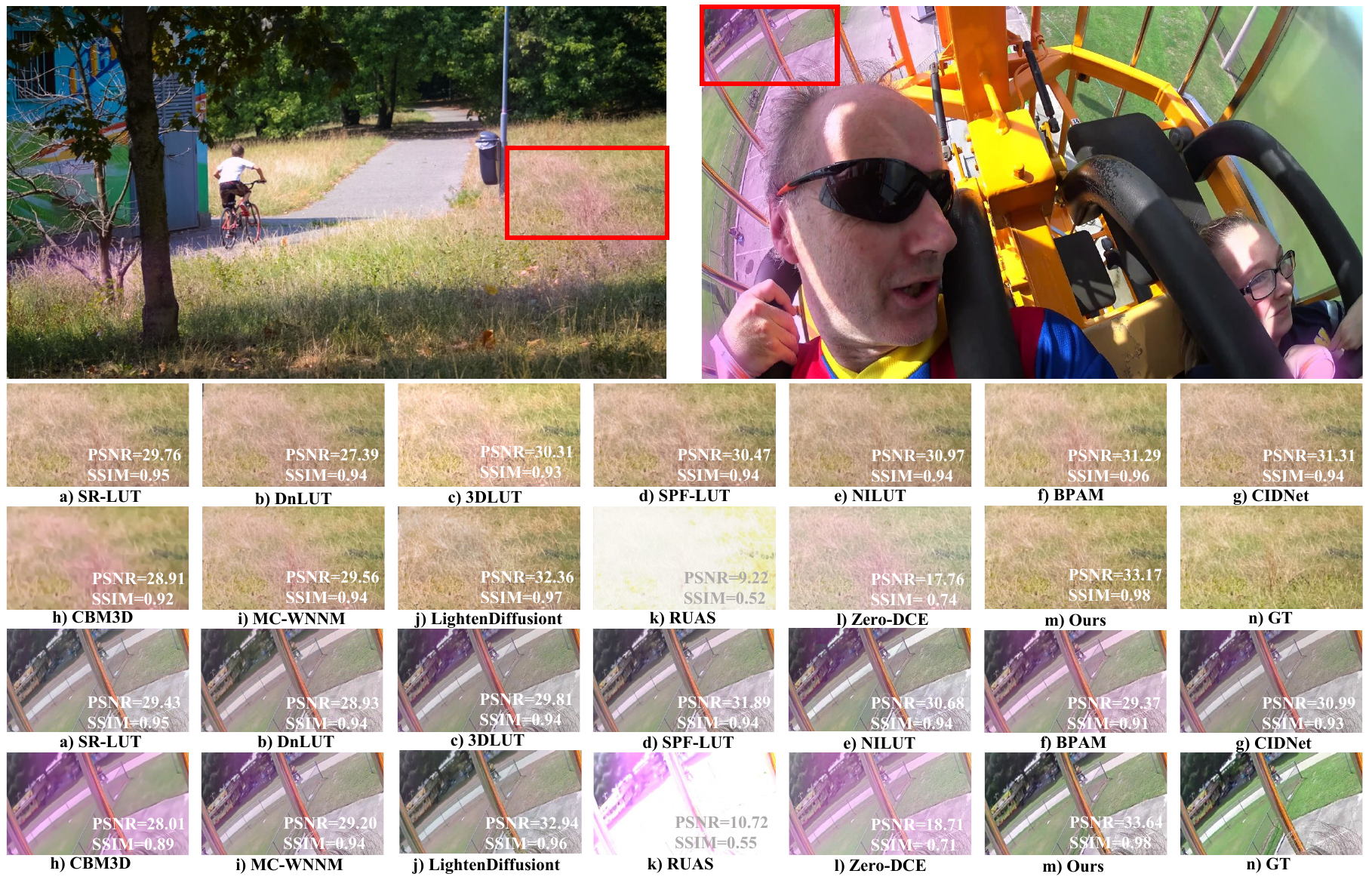} 
    \caption{Qualitative comparison with state-of-the-art methods on our PFSD dataset.}
    \label{fig:SOTA}
    \vspace{-2mm}
\end{figure*}
\section{Experiments}

\subsection{Experimental Settings}
\noindent \textbf{Datasets and Baselines. }
We conduct experiments on our new Purple Flare Synthesis Dataset (PFSD), created by applying a parametric synthesis pipeline to high-resolution frames from the DAVIS dataset. Our dataset consists of 4,987 pairs for training, 608 pairs for validation, and 618 pairs for testing.  Further details on the dataset construction and the source code can be found in the supplementary material. For a comprehensive evaluation, we compare our method against state-of-the-art (SOTA) baselines from three distinct categories: classical, LUT-based, and scene-specific restoration methods.

\begin{figure*}[t]
    \centering
    \includegraphics[width=\textwidth]{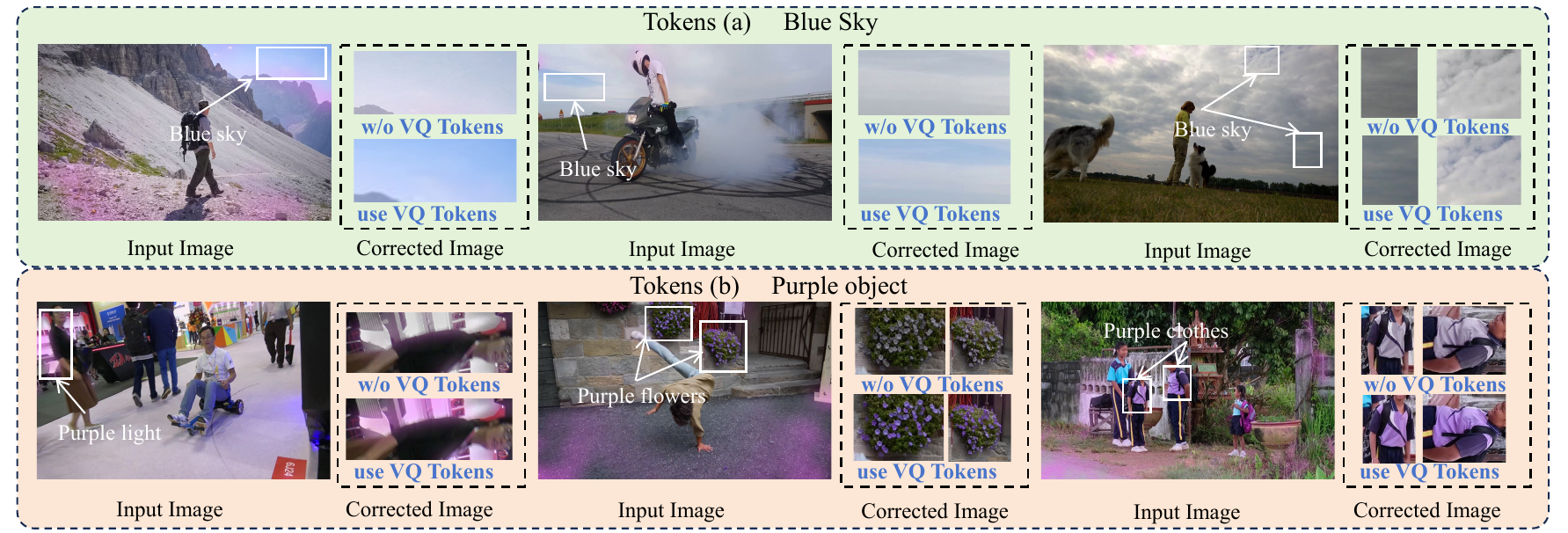} 
    \caption{Visualization of learned semantic tokens from the CAST module.}
    \label{fig:cast_viz}
\end{figure*}
\noindent \textbf{Metrics.}
To comprehensively evaluate our method, we report standard metrics: PSNR and SSIM for image fidelity, LPIPS for perceptual quality, and $\Delta E$ in the CIELAB space for color difference and efficiency measures: model parameters, FLOPs, and runtime. However, since global metrics fail to simultaneously evaluate targeted artifact removal and the preservation of clean regions, we propose a more fine-grained protocol. 
we introduce two specialized metrics: Flare Region PSNR (\textbf{PSNR-F}) to measure removal effectiveness and Non-Flare Region PSNR (\textbf{PSNR-NF}) to quantify detail preservation in unaffected areas.

Furthermore, to specifically assess the accuracy of color restoration, we propose the Hue Alignment Error (\textbf{HAE}). This perceptually-driven metric calculates the saturation-weighted circular hue difference within the flare region, reflecting that the human eye is more sensitive to hue errors in vibrant colors.
\begin{equation}
    \text{HAE} = \frac{\sum_{(\mathbf{x,y}) \in \mathbf{M}_{flare}} \Delta H(\mathbf{x,y}) \cdot \mathbf{S}_{GT}(\mathbf{x,y})}{\sum_{(\mathbf{x,y}) \in \mathbf{M}_{flare}} \mathbf{S}_{GT}(\mathbf{x,y}) + \epsilon},
\end{equation}
where $\Delta H$ is the circular hue difference. Together, PSNR-F/NF and HAE provide a comprehensive framework for evaluating this localized removal task.


\noindent \textbf{Implementation Details. }
The model was implemented in PyTorch and trained on an NVIDIA RTX 4090 (32G) via a two-stage strategy: a pre-trained CAST module (4096-entry codebook, 128-dim embedding) was loaded and frozen, then the main network trained for 100 epochs with AdamW (batch size 8, initial lr $1 \times 10^{-4}$ decayed by cosine annealing).  Inputs were resized to $256 \times 256$ with random horizontal flips and color jittering.  Composite loss weights were set to 1.0, 0.1, 2.0, and 0.1 for $\lambda_{1}, \lambda_{p}, \lambda_{f}, \lambda_{q}$.

\subsection{Comparison with State-of-the-Art Methods}

\noindent \textbf{Quantitative results.}
In the Table. \ref{tab:sota_comparison}, comprehensive quantitative comparisons on the PFSD test set demonstrate that our method achieves an HAE score of 4.10, the lowest among all methods, largely confirming its state-of-the-art performance in preserving chromatic fidelity. While general restoration methods like HVI-CIDNet perform well on global metrics such as $\Delta E$, they struggle with targeted chromatic correction. CAST-LUT consistently outperforms all baselines across standard and artifact-specific metrics, validating its overall restoration quality and perceptual similarity to the ground truth, while its lightweight design ensures efficient runtime for real-time use on resource-constrained devices.

\noindent \textbf{Qualitative results.}
Fig.~\ref{fig:SOTA} shows the visual comparisons on two challenging scenes from our test set. The first example (top row, a cyclist near grass) presents a common failure case where methods might misidentify and desaturate the naturally green-colored grass. As seen in the magnified patches, methods like d) and l) incorrectly alter the color of the grass. In contrast, our method (m) precisely removes the purple flare from the high-contrast areas while perfectly preserving the color fidelity of the grass. The second example (bottom row, a person in a glider) shows a sky region severely corrupted by purple flare. While most competing methods either fail to completely remove the purple cast (e.g., g, h) or introduce unnatural color shifts (e.g., d, i), our method successfully restores the natural sky color and cloud details without any visible artifacts, demonstrating the effectiveness of our perception-correction paradigm.

\begin{table}[t]
\centering
\resizebox{\columnwidth}{!}{%
\begin{tabular}{@{}llcccc@{}}
\toprule
\textbf{ID} & \textbf{Model Configuration} & \textbf{PSNR $\uparrow$} & \textbf{PSNR-F $\uparrow$} & \textbf{PSNR-NF $\uparrow$} & \textbf{HAE $\downarrow$} \\
\midrule
\multicolumn{6}{c}{\textit{(1) Analysis of Core Correction Space \& Mechanism}} \\
\midrule
M1 & w/o HSV (uses RGB 1D-LUTs) & 28.04 & 24.15 & 29.81 & 8.81 \\
M2 & w/o Decoupled LUTs (uses 3D RGB-LUT) & 30.05 & 25.66 & 30.95 & 7.79 \\
\midrule
\multicolumn{6}{c}{\textit{(2) Analysis of Guidance Module}} \\
\midrule
M3 & w/o CAST (simple CNN encoder) & 31.54 & 29.01 & 29.42 & 6.5 \\
M4 & w/o VQ in CAST (continuous features) & 29.50 & 28.45 & 29.49 & 6.7 \\
\midrule
\multicolumn{6}{c}{\textit{(3) Analysis of Architecture}} \\
\midrule
M5 & w/o Residual Branch & 25.82 & 26.95 & 27.37 & 5.3 \\
\midrule
\textbf{Full} & \textbf{CAST-LUT (Ours)} & \textbf{34.96} & \textbf{30.74} & \textbf{34.35} & \textbf{4.1} \\
\bottomrule
\end{tabular}%
}
\caption{Ablation study of the CAST-LUT framework.}
\vspace{-4mm}
\label{tab:ablation}
\end{table}

\subsection{Ablation study}
To validate our design choices and quantify each key component’s contribution in CAST-LUT, we conducted comprehensive ablation experiments: starting with the full model, we systematically removed or replaced individual modules and analyzed performance impacts via our metrics. Results in Table~\ref{tab:ablation} confirm the efficacy of each proposed component.


\noindent \textbf{Effectiveness of the Decoupled HSV Space.}We validate our core mechanism by testing variants that operate in RGB space. As shown in Tab.~\ref{tab:ablation}, removing decoupled HSV correction and using RGB 1D-LUTs (M1) causes a catastrophic drop in performance, with PSNR-F plummeting from 30.74 to 24.15 and HAE worsening from 4.1 to 8.81. This highlights the severe limitations of color-coupled spaces. Similarly, using a standard 3D RGB-LUT (M2) is also significantly inferior, confirming the superiority of our decoupled 1D-LUT design for precise, targeted correction.
\begin{table}[b]
\centering
\setlength{\tabcolsep}{2pt}
\begin{tabular}{@{}lcccc@{}}
\toprule
\textbf{Variant} & $N_L=1$  & $N_L=8$ & $N_L=16$ (Ours) & $N_L=32$ \\
\midrule
\textbf{PSNR $\uparrow$} & 30.11 & 31.35 & \textbf{34.96} & 32.43  \\
\textbf{HAE $\downarrow$} & 5.1 & 4.4 & \textbf{4.1} & 5.8 \\
\bottomrule
\end{tabular}
\caption{Ablation on the number of fused LUTs ($N_L$). }
\label{tab:ablation_nl}
\end{table}

\noindent\textbf{Effectiveness of the CAST Guidance.}
To show our semantic guidance's importance, we replaced CAST with a simple CNN encoder (M3) and disabled its VQ step (M4).   M3 drops PSNR-F to 29.01, M4 further to 28.45.   This confirms CAST's discrete semantic tokens offer more robust, effective guidance than simple low-level features.
Figure~\ref{fig:cast_viz} intuitively demonstrates the semantic vocabulary learned by CAST using two key examples, comparing the full model (use VQ Tokens) with a baseline model that directly uses continuous features for guidance (w/o VQ Tokens). We feature blue sky as it is a common background for flares, and its spectral proximity to purple severely tests the model's color stability. Concurrently, by visualizing a token for a purple object, we demonstrate CAST's ability to distinguish a legitimate object from a purple flare based on contextual features rather than color alone. This content-aware discrimination is precisely what enables our model to eliminate artifacts while preserving the color fidelity of all objects in the scene. 

\begin{figure}[t]
    \centering
    \includegraphics[width=\linewidth]{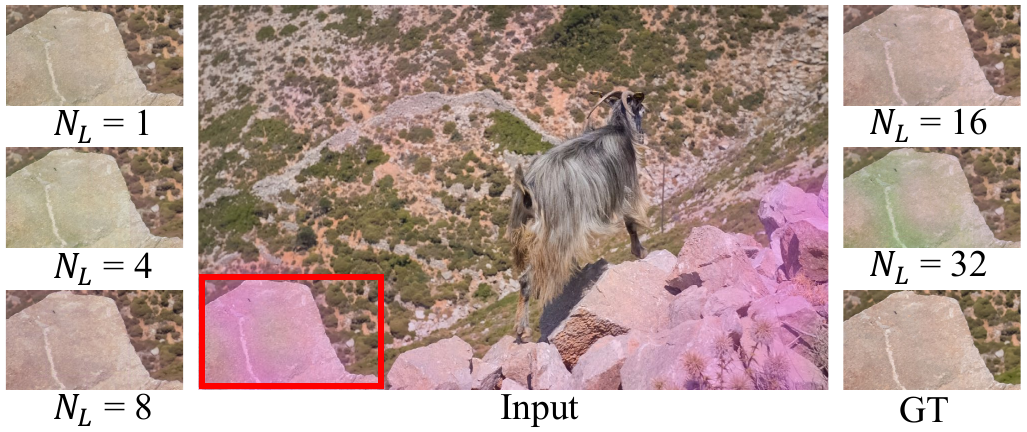}
    \caption{Visual ablation for the number of fused LUTs. 
    }
    \vspace{-3mm}
    \label{fig:lut}
\end{figure}

\noindent \textbf{Effectiveness of Architectural Choices and the number of fused LUTs ($N_L$).} 
%
Final experiments validate our key architectural and loss function choices. First, we analyze our multi-LUT fusion mechanism by ablating the fused LUT count $N_L$. Results in Table \ref{tab:ablation_nl} and Figure \ref{fig:lut} confirm performance peaks at $N_L=16$, achieving the best PSNR with effective artifact suppression.  
Other components are also critical: removing the residual branch in M5 severely degrades non-flare region detail preservation.

\begin{figure}[b]
    \centering
    \includegraphics[width=\linewidth]{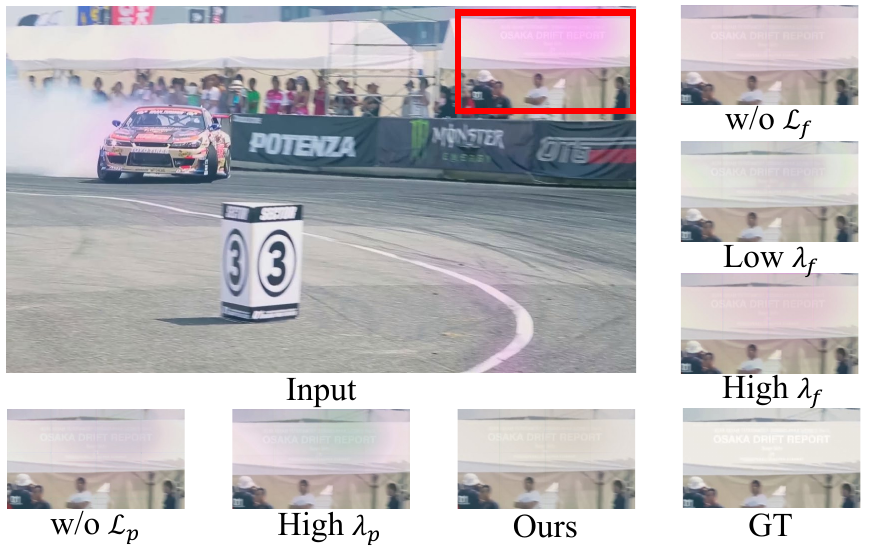}
    \caption{Visual ablation experiment of key loss terms.}
    \label{fig:loss}
\end{figure}

\noindent \textbf{Analysis of Key Loss Weights.}
As presented in Table \ref{tab:ablation_loss_weights} and Figure \ref{fig:loss}, our ablation study on the loss weights, $\lambda_f$ and $\lambda_p$, clearly demonstrates the necessity of both terms. Removing the flare suppression loss ($\mathcal{L}_f$) causes the HAE metric to worsen dramatically from 4.1 to 7.2, while removing the perceptual loss ($\mathcal{L}_p$) results in a significantly poorer LPIPS score of 0.061. Our final balanced configuration achieves the best overall performance by effectively navigating the trade-off between pixel-wise accuracy and perceptual quality.

\begin{table}[t]
\centering
\setlength{\tabcolsep}{3.5pt} 
\begin{tabular}{@{}llccccc@{}}
\toprule
\textbf{ID} & \textbf{Variant} & $\lambda_{f}$ & $\lambda_{p}$ & \textbf{PSNR $\uparrow$} & \textbf{HAE $\downarrow$} & \textbf{LPIPS $\downarrow$} \\
\midrule
I   & w/o $\mathcal{L}_{f}$   & 0.0 & 0.1 & 32.04 & 7.2 & 0.058 \\
II  & Low $\lambda_{f}$       & 0.5 & 0.1 & 32.21 & 5.8 & 0.054 \\
III & High $\lambda_{f}$      & 5.0 & 0.1 & 32.32 & 5.9 & 0.052 \\
\midrule
IV  & w/o $\mathcal{L}_{p}$   & 2.0 & 0.0 & 32.38 & 4.2 & 0.061 \\
V   & High $\lambda_{p}$      & 2.0 & 0.5 & 32.15 & 5.6 & 0.049 \\
\midrule
\multicolumn{2}{@{}l}{\textbf{Ours (Balanced)}} & \textbf{2.0} & \textbf{0.1} & \textbf{34.96} & \textbf{4.1} & \textbf{0.03} \\
\bottomrule
\end{tabular}
\caption{Ablation study on the weights of the Purple Flare Suppression Loss ($\lambda_{f}$) and the Perceptual Loss ($\lambda_{p}$). }
\label{tab:ablation_loss_weights}
\vspace{-3mm}
\end{table}


\section{Disscuss}
\begin{figure}[h]
    \centering
    \includegraphics[width=0.9\linewidth]{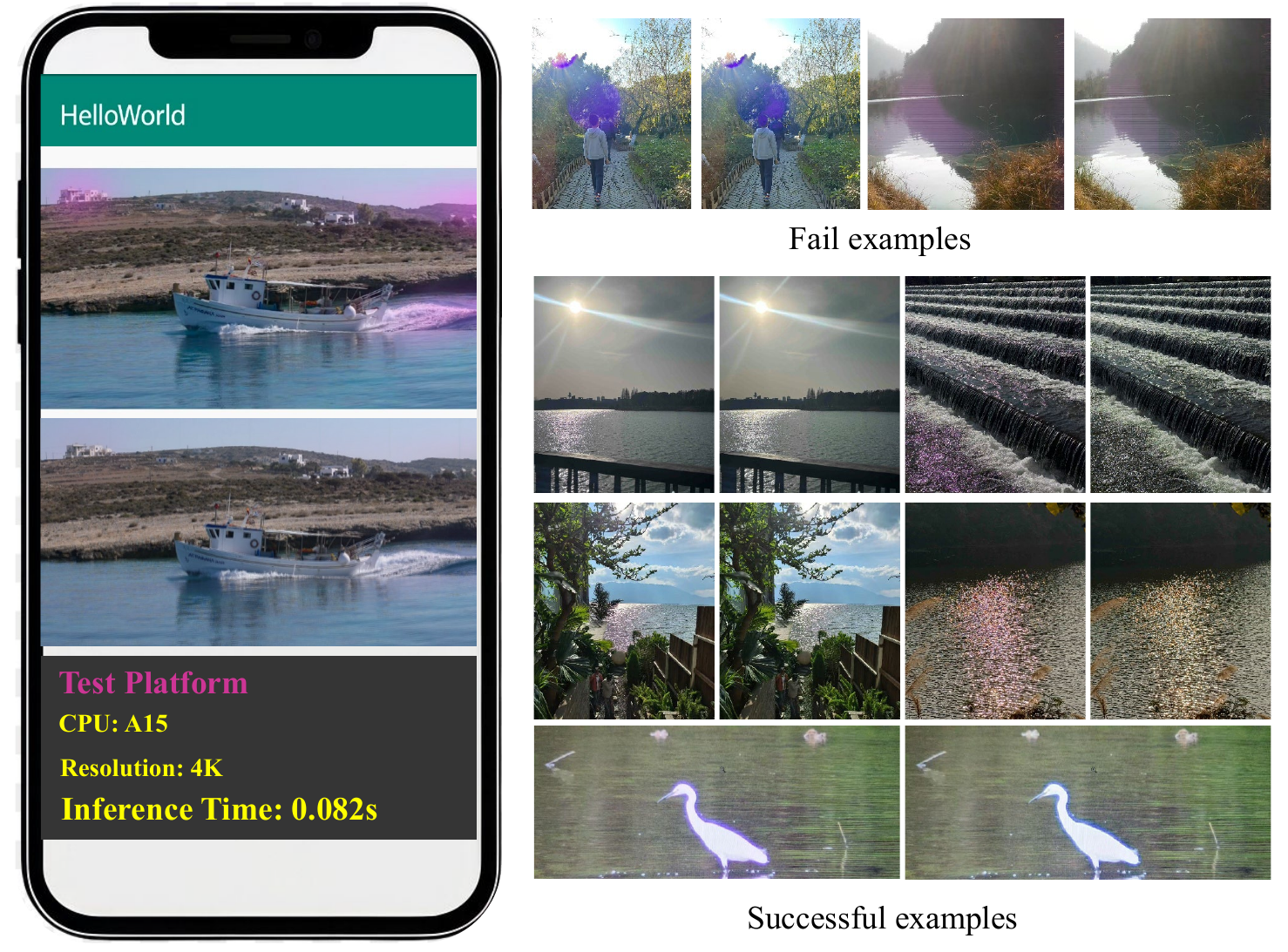}
    \caption{Sample image purple flare removing application on mobile devices based on the proposed CAST-LUT and fail and succcessful examples in real world.}
    \label{fig:enter-label}
    \vspace{-3mm}
\end{figure}

To verify practical applicability, we deployed the CAST-LUT model on mobile devices, achieving real-time performance.    Leveraging adaptive 1D HSV LUTs, our approach is far more efficient than methods relying on large 3D LUTs or complex CNNs, while preserving fine details in unaffected areas through pixel-wise correction.    Specifically, tests on the iPhone 14 recorded an 82ms execution time for 4K processing, demonstrating efficient hardware-software compatibility. Fig.~\ref{fig:enter-label} shows the mobile application's interface and some failure and successful cases in real world. Supplementary material includes leading methods’ real-world results for broader comparison.

\section{Conclusion}
We propose CAST-LUT, a purple flare removal framework operating in decoupled HSV space to address traditional methods’ poor generalization and RGB-to-HSV conversion noise.  Its core is CAST, which encodes purple flare artifacts into stable high-level semantic tokens.  This avoids direct conversion noise while guiding the generation of independent 1D-LUTs for precise color correction.  Extensive experiments on our large-scale PFSD dataset, using metrics like PSNR-F/NF and HAE, confirm its superior artifact removal and detail preservation performance.

\section{Acknowledgments}
This work was supported by the National Natural Science Foundation of China Project (No. 62172265) and Shandong Provincial Natural Science Foundation (ZR2025MS1025, ZR2025MS1036).

\bibliography{aaai2026}

\makeatletter
\@ifundefined{isChecklistMainFile}{
  \newif\ifreproStandalone
  \reproStandalonetrue
}{
  \newif\ifreproStandalone
  \reproStandalonefalse
}
\makeatother

\ifreproStandalone

\fi
\setlength{\leftmargini}{20pt}
\makeatletter\def\@listi{\leftmargin\leftmargini \topsep .5em \parsep .5em \itemsep .5em}
\def\@listii{\leftmargin\leftmarginii \labelwidth\leftmarginii \advance\labelwidth-\labelsep \topsep .4em \parsep .4em \itemsep .4em}
\def\@listiii{\leftmargin\leftmarginiii \labelwidth\leftmarginiii \advance\labelwidth-\labelsep \topsep .4em \parsep .4em \itemsep .4em}\makeatother

\setcounter{secnumdepth}{0}
\renewcommand\thesubsection{\arabic{subsection}}
\renewcommand\labelenumi{\thesubsection.\arabic{enumi}}

\newcounter{checksubsection}
\newcounter{checkitem}[checksubsection]

\newcommand{\checksubsection}[1]{%
  \refstepcounter{checksubsection}%
  \paragraph{\arabic{checksubsection}. #1}%
  \setcounter{checkitem}{0}%
}

\newcommand{\checkitem}{%
  \refstepcounter{checkitem}%
  \item[\arabic{checksubsection}.\arabic{checkitem}.]%
}
\newcommand{\question}[2]{\normalcolor\checkitem #1 #2 \color{blue}}
\newcommand{\ifyespoints}[1]{\makebox[0pt][l]{\hspace{-15pt}\normalcolor #1}}

\clearpage
\setcounter{section}{0}

\section{Supplementary Material}

This supplementary material provides additional details to support our paper. Section A elaborates on the network architecture, including the specific implementations of the LUTGenerator, and WeightGenerator. Section B describes our synthetic dataset (PFSD), which covers the parameterized flare synthesis process and key hyperparameters. Section C presents further ablation analysis, offering additional studies on the impact of VQ codebook size and encoder/decoder depth on model performance and efficiency. Section D showcases extensive qualitative results, with visual comparisons against state-of-the-art methods on complex scenes, night scenes from the synthetic PFSD dataset, and various real-world images.

\section*{A. HSV-LUT Color Correction Module and Implementation}\label{sec:1}

\subsection*{A.1. LUTGenerator Architecture.}
The LUTGenerator is implemented as a Multi-Layer Perceptron (MLP) that maps the input feature vector $\mathbf{f}_{\text{token}}$ to a flat vector containing all necessary parameters for the 1D-LUTs. Its architecture is as follows:

\noindent \textbf{Layer 1.} A linear layer that maps the input dimension from hidden\_dim to hidden\_dim.

\noindent \textbf{Activation.} A GELU activation function (\texttt{nn.GELU}).

\noindent \textbf{Layer 2 (Output).} A final linear layer that maps from hidden\_dim to the total number of LUT parameters. The output dimension is calculated as:
\begin{equation}
    \mathbf{D}_\text{out} = N_{L} \times \mathbf{C}_\text{HSV} \times \mathbf{S}_\text{LUT},
\end{equation}
 where $N_L$ is the number of LUT sets, $\mathbf{C}_\text{HSV}=3$ for the H, S, and V channels, 
and the size of each individual 1D-LUT is determined by $\mathbf{S}_\text{LUT}$ control points.

The resulting flat vector of dimension $D_{out}$ is then reshaped into a structured tensor $\mathbf{P} \in \mathbb{R}^{B \times N_L \times \mathbf{C}_\text{HSV} \times \mathbf{S}_\text{LUT}}$, where $B$ is the batch size.

\subsection*{A.2. WeightGenerator Architecture.}
Running in parallel to the LUTGenerator, the WeightGenerator is another MLP designed to predict the fusion weights for each of the $N_L$ LUT sets. Its architecture is as follows:

\noindent \textbf{Layer 1.} A linear layer that maps the input dimension from hidden\_dim to a smaller intermediate dimension hidden\_dim / 4. Given the input feature vector $\mathbf{f}_{\text{token}}$, this operation produces a pre-activation vector $\mathbf{h'}$:
\begin{equation}
    \mathbf{h'} = \mathbf{f}_{\text{token}} W_1^T + b_1
\end{equation}

\noindent \textbf{Activation.} A GELU activation function (\texttt{nn.GELU}) is applied to the pre-activation vector $\mathbf{h'}$ to produce the activated intermediate representation $\mathbf{h}$:
\begin{equation}
    \mathbf{h} = \text{GELU}(\mathbf{h'})
\end{equation}

\noindent \textbf{Layer 2 (Output).} A final linear layer maps the intermediate representation $\mathbf{h}$ to a vector of $N_L$ raw scores (logits), denoted as $\mathbf{z}$:
\begin{equation}
    \mathbf{z} = \mathbf{h} W_2^T + b_2
\end{equation}

A Softmax function is subsequently applied to these logits to ensure the final weights $\mathbf{W}$ sum to one, forming a valid probability distribution:
\begin{equation}
    \mathbf{W}_i = \text{Softmax}(\mathbf{z})_i = \frac{\exp(\mathbf{z}_i)}{\sum_{j=1}^{N_L} \exp(\mathbf{z}_j)}
\end{equation}

\section*{B.Purple Flare Synthesis Dataset (PFSD)}\label{sec:2}


Our Purple Flare Synthesis Dataset (PFSD) is the first large-scale, paired dataset constructed for this task. We first select diverse, clean image frames from the high-resolution video dataset DAVIS as our ground truth, denoted as \( \mathbf{I}_\text{GT} \). We then generate purple flare input images \( \mathbf{I}_\text{flare} \) using a physically motivated function, designed to mimic the characteristic appearance of purple flare in high-luminance regions and at object contour edges. This process is governed by a set of key hyperparameters, including the highlight percentile $\rho_h$, gradient threshold $\tau_g$, edge width $w_{edge}$, blend strength $\alpha_s$, and radial gamma $\gamma$.
The specific generation process is as follows: 

\subsubsection*{Dynamic Highlight Detection}
To identify potential flare regions, we first calculate a dynamic highlight threshold, $\tau_h$, defined as the $\rho_h$-th percentile of the grayscale input image $\mathbf{I}_\text{gray}$. We then generate a binary highlight mask $\mathbf{M}_\text{bright}$ as follows:
\begin{equation}
    \mathbf{M}_\text{bright} = (\mathbf{I}_\text{gray} > \tau_h).
\end{equation}

\begin{table*}[ht!]
\centering
\caption{Key parameters for the synthetic purple flare generation pipeline. Default values are those used to generate our dataset.}
\label{tab:flare_params}
\begin{tabular}{@{}llc@{}}
\toprule
\textbf{Parameter} & \textbf{Description} & \textbf{Value} \\
\midrule
\texttt{highlight\_pct} & Percentile for the dynamic highlight threshold. & 99.0 \\
\texttt{grad\_thresh} & Gradient magnitude threshold for the Sobel edge detector. & 25 \\
\texttt{edge\_width} & Width (in pixels) of the flare band. & 80 \\
\texttt{strength} & Global blending strength of the purple overlay. & 0.7 \\
\texttt{gamma} & Exponent for the radial mask to intensify the corner effect. & 2.2 \\
\bottomrule
\end{tabular}
\end{table*}

\subsubsection*{Edge Candidate Identification}
We use a Sobel operator to compute the gradient magnitude $G$ of the grayscale image. An edge mask $\mathbf{M}_\text{edge}$, is formed by thresholding the gradient magnitude with $\tau_g$:
\begin{equation}
    G = \sqrt{(\text{Sobel}_x * \mathbf{I}_\text{gray})^2 + (\text{Sobel}_y * \mathbf{I}_\text{gray})^2},
\end{equation}
where $\quad \mathbf{M}_\text{edge} = (G > \tau_g)$.

\subsubsection*{Flare Mask Generation}
The initial region for the purple flare is identified as the intersection of highlight and edge regions, creating a candidate mask $\hat{\mathbf{M}}_\text{flare}$:
\begin{equation}
    \hat{\mathbf{M}}_\text{flare} = \mathbf{M}_\text{bright} \cap \mathbf{M}_\text{edge}.
\end{equation}

\subsubsection*{Flare Simulation}
To simulate the width and diffusion properties of flares, we expand the candidate mask $\hat{\mathbf{M}}_\text{flare}$ by applying morphological dilation with a structural element $K_e$, whose size is determined by the hyperparameter $w_{edge}$. 
\begin{equation}
    \mathbf{M}_\text{dilated} = \hat{\mathbf{M}}_\text{flare} \oplus K_e,
\end{equation}
where $\oplus$ denotes morphological dilation. 

$\mathbf{M}_\text{dilated}$ is convolved with a Gaussian kernel $G_\sigma$ to create soft edges, generating the final flare band mask. 
\begin{equation}
    \mathbf{M}_\text{flare} = \mathbf{M}_\text{dilated} * G_\sigma,
\end{equation}
where the standard deviation $\sigma$ is proportional to $w_{\text{edge}}$. 

\subsubsection*{Spatially-Varying Alpha Mask}
To model the influence of lens geometry, where flares typically intensify toward the corners, we generate a radial attenuation mask \( \mathbf{R} \). 
\begin{equation}
    \mathbf{R}(x,y) = \left( \frac{\text{dist}\bigl( (x,y), \text{center} \bigr)}{\dist_{\text{max}}} \right)^\gamma,
\end{equation}
This mask $\mathbf{R}$ is combined with the normalized flare band $\mathbf{M}_\text{flare}$ and the blend strength $\alpha_s$ to create the final spatially-varying alpha mask $\alpha$:
\begin{equation}
    \alpha = \frac{\mathbf{M}_\text{flare}}{\max(\mathbf{M}_\text{flare})} \cdot \mathbf{R} \cdot \alpha_s.
\end{equation}

\subsubsection*{Alpha Blending}
The final purple flare image \( \mathbf{I}_{\text{flare}} \) is generated by alpha blending the ground truth image \( \mathbf{I}_{\text{GT}} \) with a predefined purple color \( C_p \) using the generated alpha mask.
\begin{equation}
    \mathbf{I}_\text{flare} = \mathbf{I}_\text{GT} \cdot (1 - \alpha) + C_p \cdot \alpha
\end{equation}

The key parameters controlling the flare's appearance are detailed in Table~\ref{tab:flare_params}. The entire synthesis process is detailed in Algorithm~\ref{alg:synthesis}. After generation, the dataset is divided into training, validation, and test sets. To prevent data leakage where frames from the same video sequence might appear in both training and evaluation splits, the division is performed at the scene level. The list of all scenes is randomly shuffled, and entire scenes are then assigned to the training, test, and validation sets, with proportions of 80\%, 10\%, and 10\%, respectively.

Finally, the candidate mask $\hat{\mathbf{M}}_\text{flare}$ is directly used as the ground-truth mask $\mathbf{M}_\text{GT}$ for metric calculation. We select this explicit mask instead of using the diffuse final flare band mask $\mathbf{M}_\text{flare}$ for visual blending. This is because $\hat{\mathbf{M}}_\text{flare}$ enables accurate measurement of the model's ability to restore core-damaged regions, avoiding penalties for failing to precisely reconstruct the feathered halo details of simulated flares. Let $\mathbf{I}_\text{out}$ be the model's output image; the PSNR calculations for the flare region (PSNR-f) and non-flare region (PSNR-nf) are as follows: 
\begin{equation}
\text{PSNR-F} = 10 \log_{10} \left( \frac{\text{MAX}_I^2 \cdot \sum \mathbf{M}_\text{GT}}{\sum (\mathbf{I}_\text{out} - \mathbf{I}_\text{GT})^2 \odot \mathbf{M}_\text{GT}} \right),
\end{equation}
\begin{equation}
\text{PSNR-NF} = 10 \log_{10} \left( \frac{\text{MAX}_I^2 \cdot \sum (1-\mathbf{M}_\text{GT})}{\sum (\mathbf{I}_\text{out} - \mathbf{I}_\text{GT})^2 \odot (1-\mathbf{M}_\text{GT})} \right),
\end{equation}
where $\odot$ is element-wise multiplication.

To specifically evaluate the chromatic accuracy of the restoration, we designed the Hue Alignment Error (HAE), a metric more perceptually relevant than standard signal-based errors like PSNR. HAE is defined as the saturation-weighted average of the circular hue difference between the model's output $\mathbf{I}_\text{out}$ and the ground-truth $\mathbf{I}_\text{GT}$, calculated exclusively within the regions of the original input image $\mathbf{I}_\text{in}$ affected by purple flare. A lower HAE score signifies a more accurate color restoration.

\begin{algorithm}
\caption{Parametric Purple Flare Synthesis}
\label{alg:synthesis}
\begin{algorithmic}[l]
\Statex \textbf{Input:} Ground truth image $\mathbf{I}_\text{GT}$.
\Statex \textbf{Parameters:} Highlight percentile $\rho_h$, gradient threshold $\tau_g$, edge width $w_{edge}$, blend strength $\alpha_s$, radial gamma $\gamma$.
\Statex \textbf{Output:} Purple flare image $\mathbf{I}_\text{flare}$ and ground-truth mask $\mathbf{M}_\text{GT}$.
\State $\mathbf{I}_\text{gray} \leftarrow \text{Grayscale}(\mathbf{I}_\text{GT})$
\State $\tau_h \leftarrow \text{Percentile}(\mathbf{I}_\text{gray}, \rho_h)$
\State $\mathbf{M}_\text{bright} \leftarrow (\mathbf{I}_\text{gray} > \tau_h)$
\If{$\text{sum}(\mathbf{M}_\text{bright}) = 0$} \Return None, None \EndIf
\State $G \leftarrow \text{SobelMagnitude}(\mathbf{I}_\text{gray})$
\State $\mathbf{M}_\text{edge} \leftarrow (G > \tau_g)$
\State $\hat{\mathbf{M}}_\text{flare} \leftarrow \mathbf{M}_\text{bright} \cap \mathbf{M}_\text{edge}$
\If{$\text{sum}(\hat{\mathbf{M}}_\text{flare}) = 0$} \Return None, None \EndIf
\State $\mathbf{M}_\text{GT} \leftarrow \hat{\mathbf{M}}_\text{flare}$
\State $K_e \leftarrow \text{GetStructuringElement}(\text{shape=ELLIPSE, size}=w_{edge})$
\State $\mathbf{M}_\text{dilated} \leftarrow \text{Dilate}(\hat{\mathbf{M}}_\text{flare}, K_e)$
\State $\mathbf{M}_\text{flare} \leftarrow \text{GaussianBlur}(\mathbf{M}_\text{dilated}, \sigma=0.6 \cdot w_{edge})$
\State $R \leftarrow \text{RadialFalloffMask}(\text{shape of } \mathbf{I}_\text{GT}, \gamma)$
\State $\alpha \leftarrow \frac{\mathbf{M}_\text{flare}}{\max(\mathbf{M}_\text{flare})} \cdot \mathbf{R} \cdot \alpha_s$
\State Let $C_p$ be the target purple color (e.g., [255, 100, 255] in BGR)
\State $\mathbf{I}_\text{flare} \leftarrow \mathbf{I}_\text{GT} \cdot (1 - \alpha) + C_p \cdot \alpha$
\State \Return $\text{clip}(\mathbf{I}_\text{flare}, 0, 255)$, $\mathbf{M}_\text{GT}$
\end{algorithmic}
\end{algorithm}

A binary flare mask $\mathbf{M}_\text{flare}$ is generated from the input image to precisely identify the location of purple halos. This mask is used to locate pixels that satisfy two conditions: being on high-contrast edges detected by the Sobel filter, while exhibiting hues in the purple spectrum (e.g., 260°-340°) and significant saturation in the HSV color space. The specific calculation process is as follows: first, both the output image and the ground-truth image are converted to the HSV color space to separate hue, saturation, and brightness information. Then, we extract the hue channel $\mathbf{H}_\text{out}$ of the output image, as well as the hue $\mathbf{H}_\text{GT}$ and saturation $\mathbf{S}_\text{GT}$ channels of the ground-truth image. Within the region defined by the mask, the circular hue difference is calculated pixel by pixel.
\begin{equation}
\Delta H = \min(|\mathbf{H}_\text{out} - \mathbf{H}_\text{GT}|, 360 - |\mathbf{H}_\text{out} - \mathbf{H}_\text{GT}|), 
\end{equation}
which correctly measures the shortest distance on the 360-degree hue circle. To align with human perception, where hue errors are more visible in vibrant colors, each $\Delta H$ value is then weighted by the saturation of the corresponding pixel in the ground-truth image, $\mathbf{S}_{GT}$. The final HAE score is the average of these saturation-weighted errors, defined as:
\begin{equation}
    \text{HAE} = \frac{\sum_{(\mathbf{x,y}) \in \mathbf{M}_\text{flare}} \Delta \mathbf{H}(\mathbf{x,y}) \cdot \mathbf{S}_\text{GT}(\mathbf{x,y})}{\sum_{(\mathbf{x,y}) \in \mathbf{M}_\text{flare}} \mathbf{S}_\text{GT}(\mathbf{x,y}) + \epsilon},
\end{equation}
where $\epsilon$ is a small constant to prevent division by zero.


\section*{C. Additional Ablation Analysis}\label{sec:3}

\subsection*{C.1. Impact of Codebook Size in the VQ Module}

The VQ layer is the core of the CAST module, and its codebook size directly affects the expressive power of the semantic tokens. In Table~\ref{tab:codebook_ablation_en} and Figure~\ref{tab:codebook_ablation_en}, we study the impact of different codebook sizes on model performance. The results show that a codebook that is too small (e.g., 1024) cannot learn sufficiently rich semantic patterns, leading to a performance drop. Conversely, an overly large codebook (e.g., 8192) increases model parameters and the risk of overfitting without bringing significant performance gains. A codebook size of 4096 achieves the best balance between performance and model complexity.

\begin{table}[h!]
\centering
\caption{Visualization of Ablation study on different codebook sizes.}
\label{tab:codebook_ablation_en}
    \begin{tabular}{cccc}
        \toprule
        \textbf{Codebook Size} & \textbf{FLOPs (G)  $\downarrow$} & \textbf{PSNR $\uparrow$} & \textbf{HAE $\downarrow$} \\
        \midrule
        1024 & 19.45 & 33.52 & 6.83 \\
        2048 & 21.69 & 34.15 & 5.71 \\
        \textbf{4096 (Ours)} & \textbf{23.32} & 34.96 & \textbf{4.10} \\
        8192 & 26.05 & \textbf{35.03} & 4.15 \\
        \bottomrule
    \end{tabular}
\end{table}

\subsection*{C.2. Impact of Encoder/Decoder Depth}

The depth of the encoder and decoder in the CAST module determines the hierarchy of feature extraction. The results show in Table~\ref{tab:depth_ablation_en} and visualized in Figure~\ref{fig:depth_ablation_en}. The experiments prove that a network that is too shallow (2 layers) cannot extract robust high-level features, resulting in poor performance. While deepening the network to 6 layers brings slight performance improvements, it also significantly increases the computational cost. Therefore, a 4-layer depth is the optimal configuration we chose, striking a good balance between feature extraction capability and model efficiency.

\begin{figure}
    \centering
    \includegraphics[width=1\linewidth]{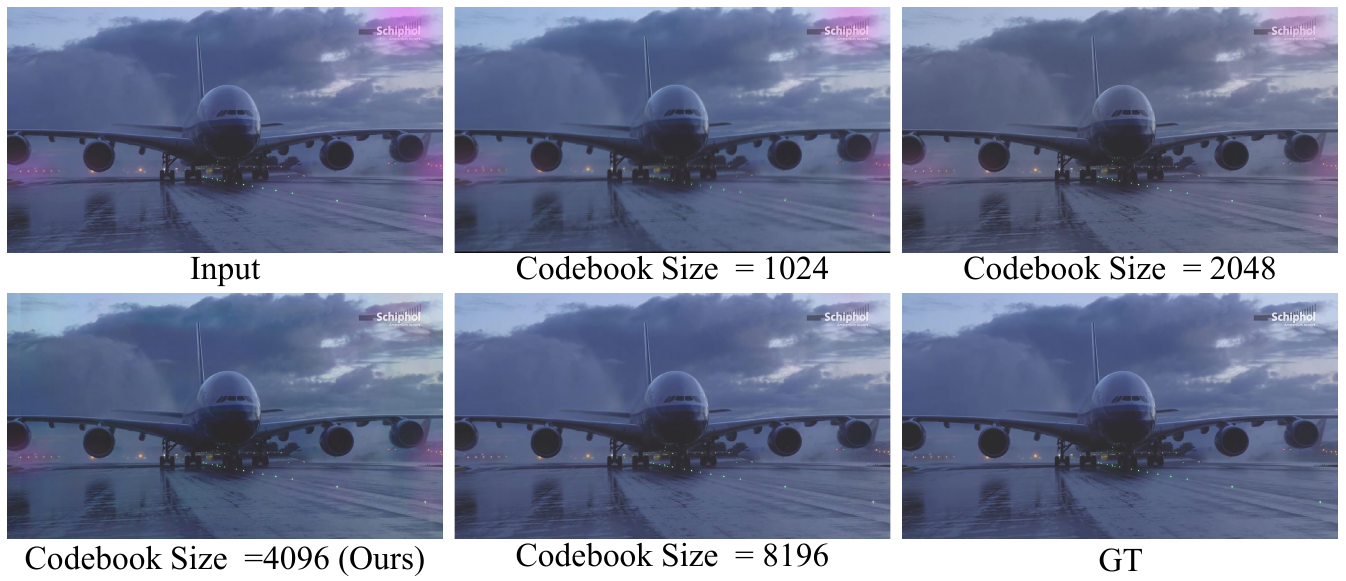}
    \caption{Ablation study on different codebook sizes.}
    \label{fig:codebook_ablation_en}
\end{figure}

\begin{figure}
    \centering
    \includegraphics[width=1\linewidth]{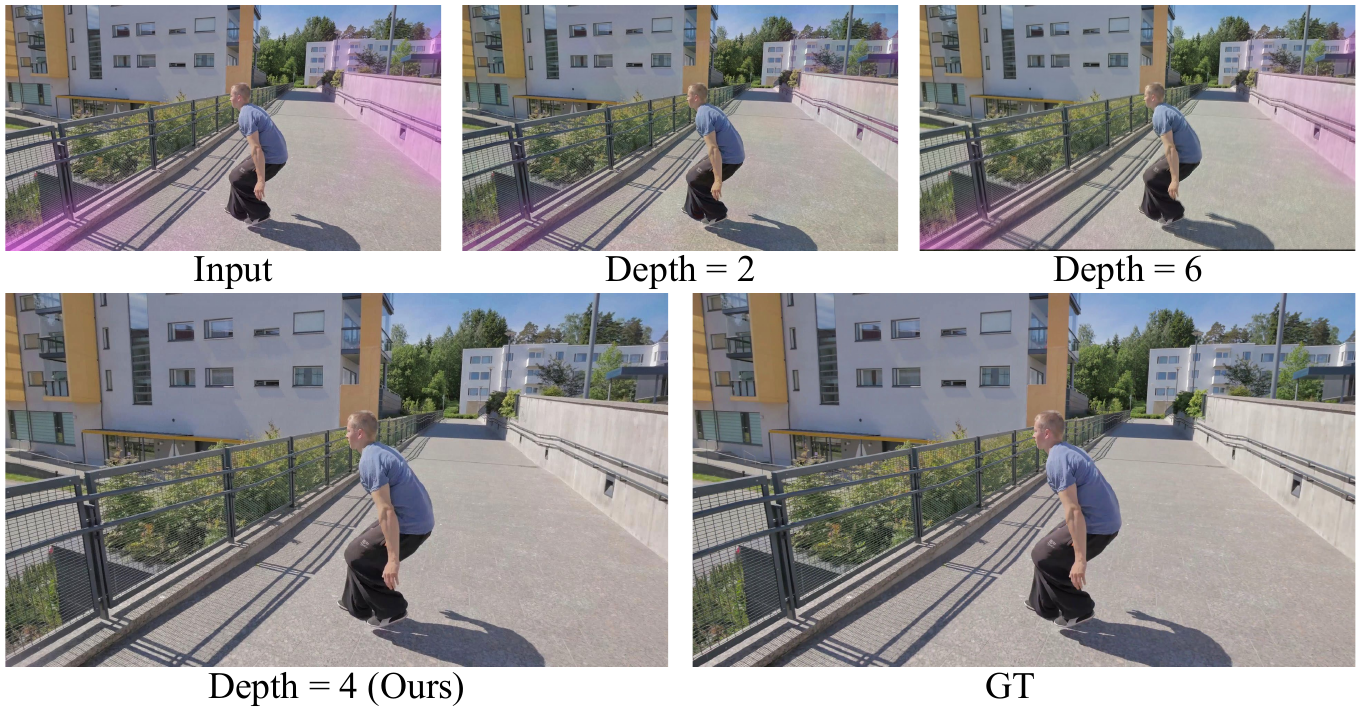}
    \caption{Visualization of the Ablation Study on Encoder/Decoder Depth.}
    \label{fig:depth_ablation_en}
\end{figure}

\begin{table}[h!]
    \centering
    \caption{Ablation study on encoder/decoder depth.}
    \label{tab:depth_ablation_en}
    \begin{tabular}{ccc}
        \toprule
        \textbf{Depth (Conv Layers)} & \textbf{PSNR $\uparrow$} & \textbf{HAE $\downarrow$} \\
        \midrule
        2 & 32.89 & 5.31 \\
        \textbf{4 (Ours)} & \textbf{34.96} & \textbf{4.10} \\
        6 & 34.91 & 4.19 \\
        \bottomrule
    \end{tabular}
\end{table}

\section*{D. More Qualitative Results}\label{sec:4}

In this section, we present extensive qualitative results to intuitively demonstrate the robustness and superiority of our proposed method. The comparisons are divided into two categories: results from complex daylight scenes and challenging night scenes in the PFSD dataset, as well as results from real-world images used to test generalization ability. 

\subsection*{D.1. Additional Qualitative Comparisons on the PFSD Dataset}

\begin{figure*}
    \centering
    \includegraphics[width=1\linewidth]{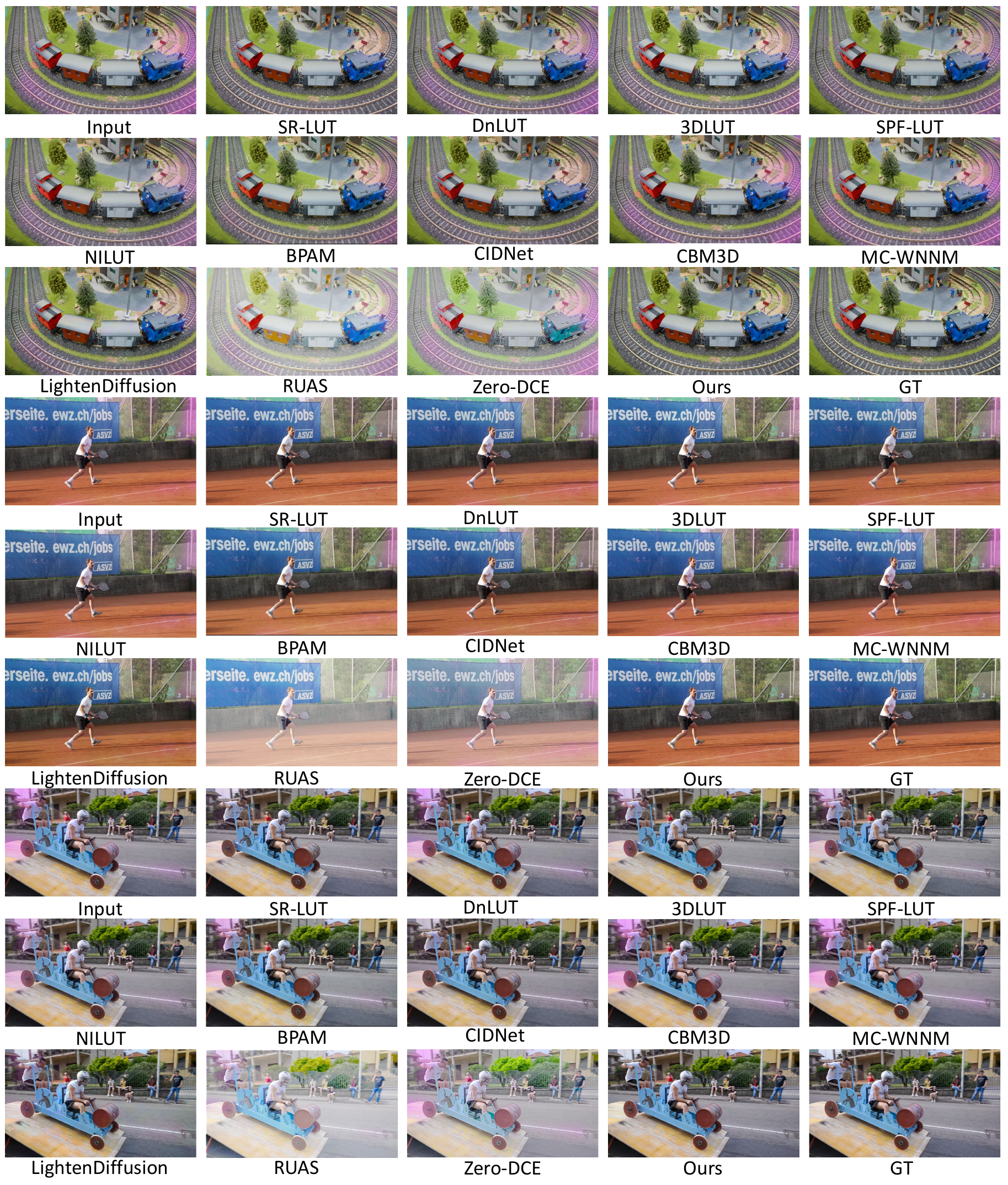}
    \caption{Visual comparison of scenes with complex textures from the PFSD dataset.. While removing purple flares, our method accurately preserves the original textures and gloss. }
    \label{fig:1}
\end{figure*}

\begin{figure*}
    \centering
    \includegraphics[width=1\linewidth]{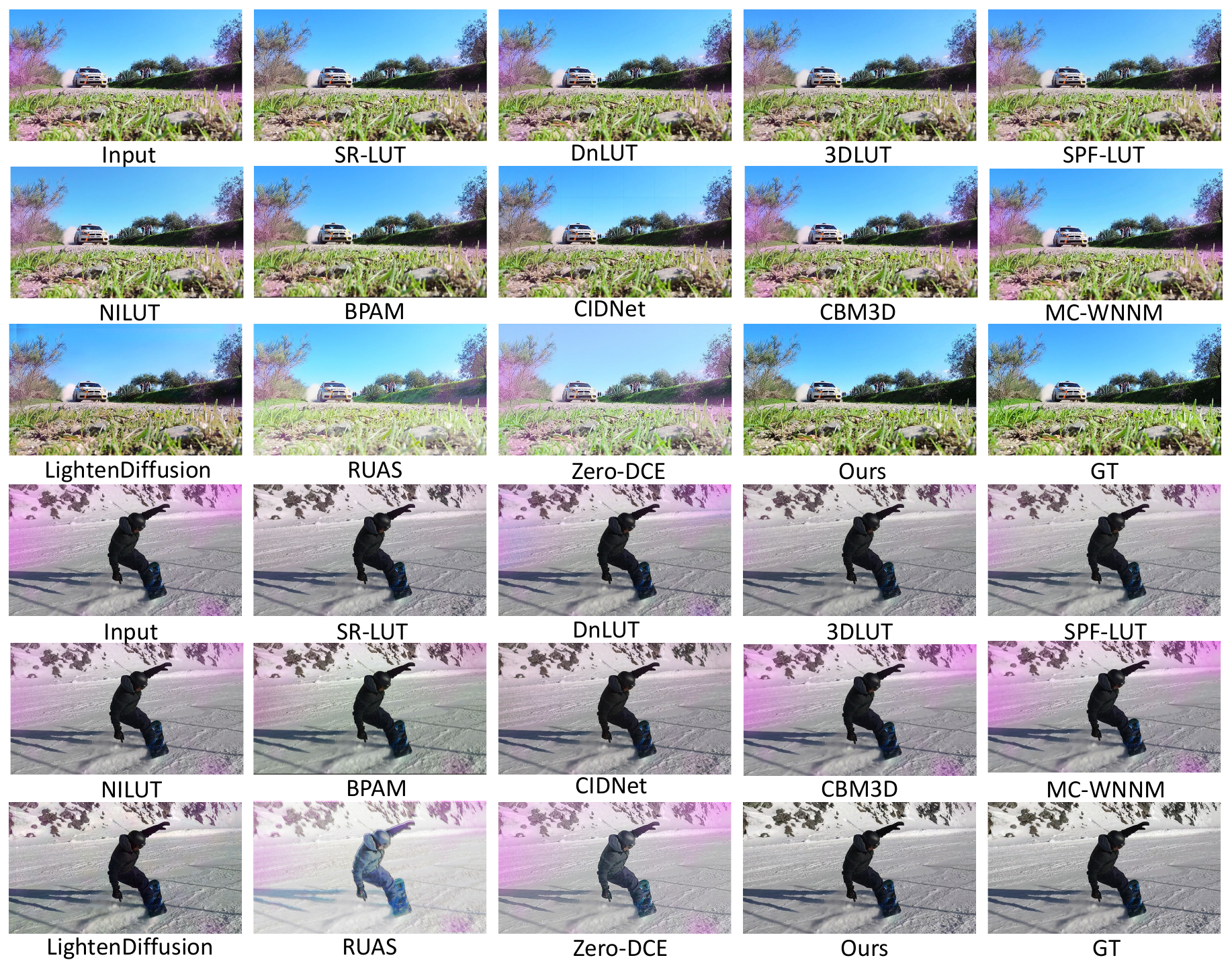}
    \caption{Visual comparison on backlit scenes from the PFSD dataset.. Our method excels at handling the severe purple flare at the boundary between the sky and trees, restoring a natural color transition.}
    \label{fig:2}
\end{figure*}

\begin{figure*}
    \centering
    \includegraphics[width=1\linewidth]{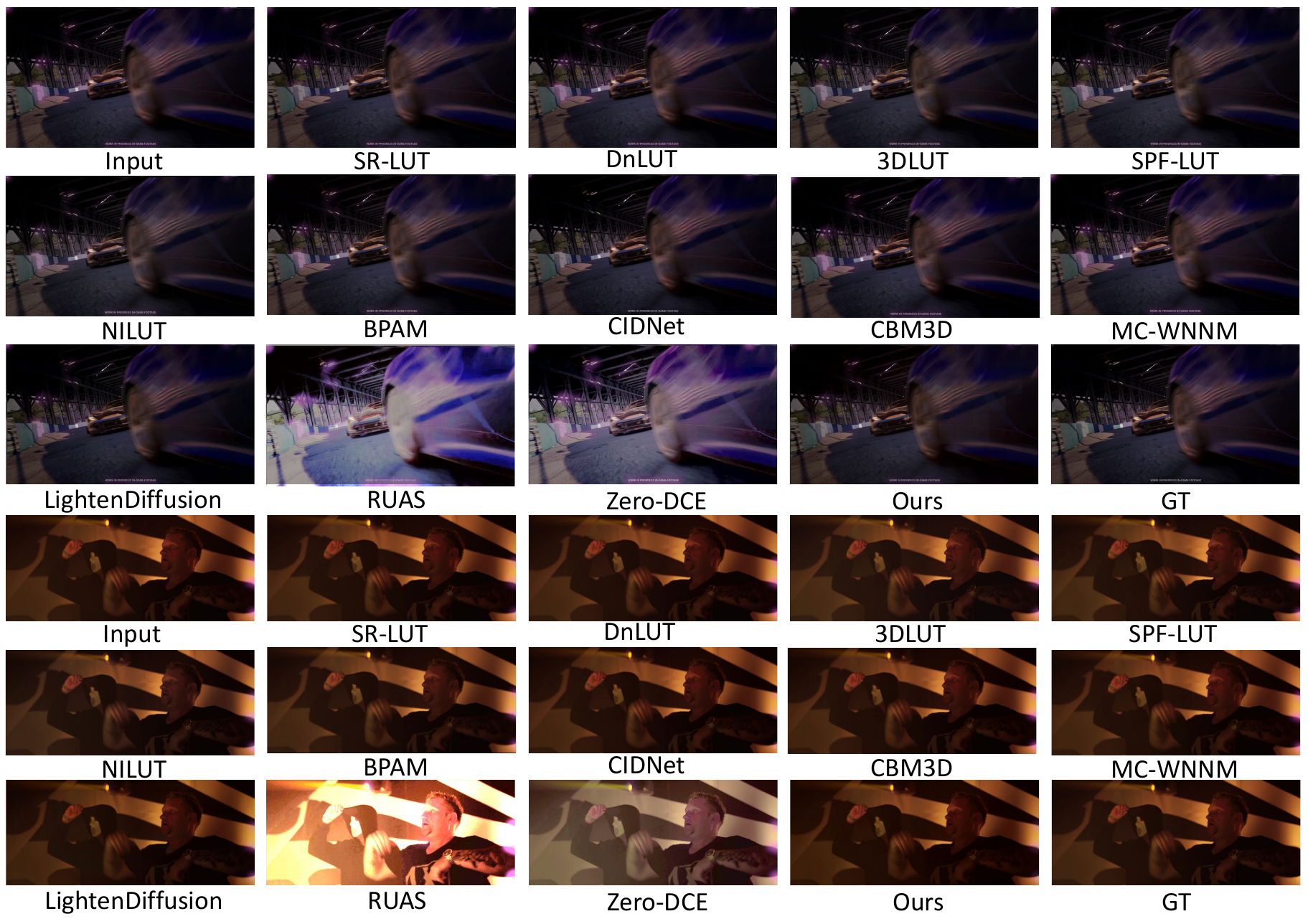}
    \caption{Visual comparison of all methods on a challenging night scene from the PFSD dataset. Our method effectively removes the purple flare around the lights while preserving details in the dark areas.}

    \label{fig:3}
\end{figure*}

To further demonstrate the comprehensive performance of the proposed method, additional comparisons with state-of-the-art approaches were conducted across various scenarios in the PFSD test set. These scenarios range from daylight scenes with severe backlighting and complex textures (Figs. \ref{fig:1} and \ref{fig:2}) to challenging night scenes (Figs. \ref{fig:3}).  

The results highlight the superiority of our method. It not only removes purple flares more effectively but also excels in preserving details and maintaining color fidelity in non-flare regions. In night scene scenarios, while other methods often leave residual artifacts or introduce unnatural color changes, the proposed method successfully corrects the images.

\subsection*{D.2. Qualitative Comparisons on Real-World Images}

\begin{figure*}
    \centering
    \includegraphics[width=1\linewidth]{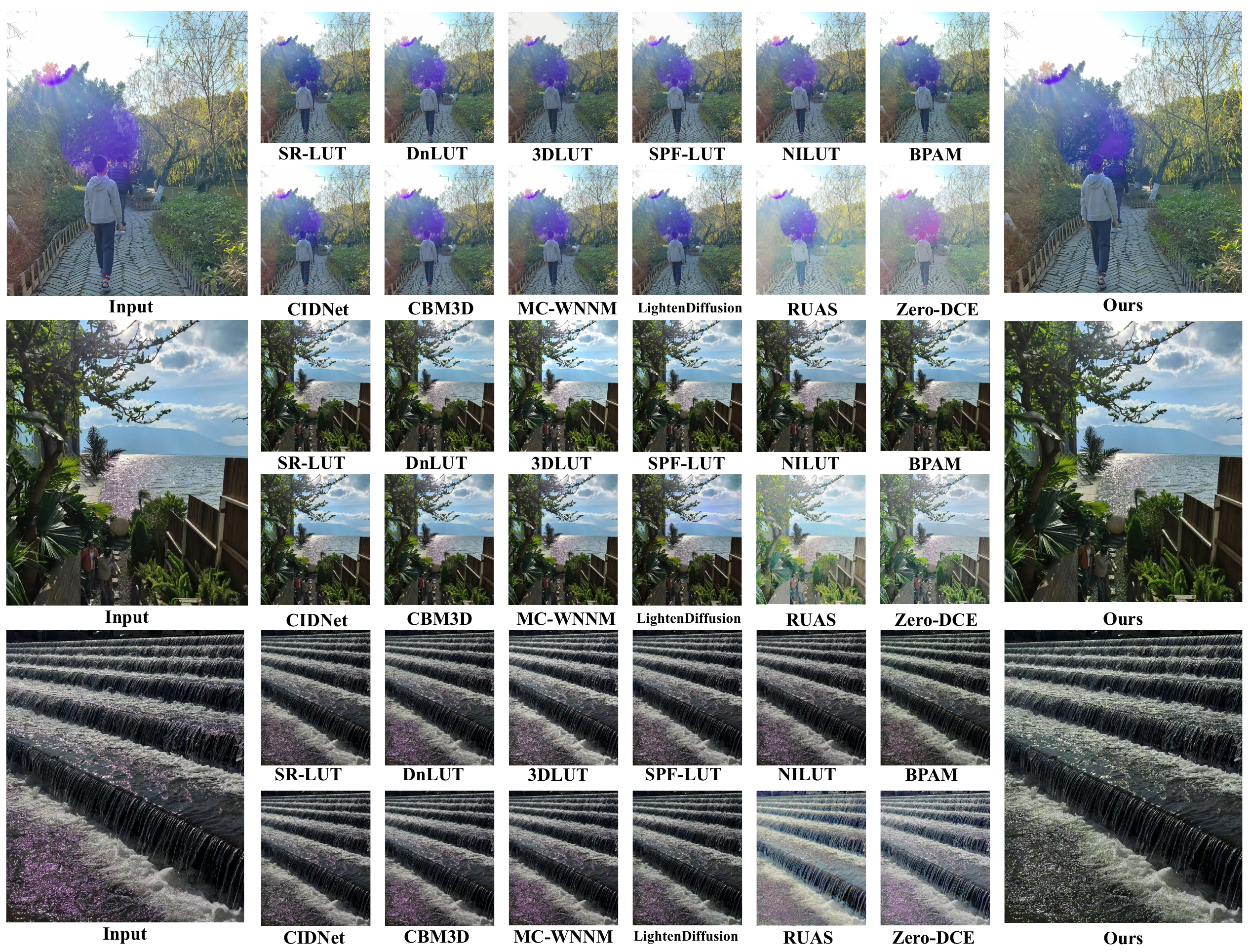}
    \caption{Visual comparison of all methods on a real-world photograph with purple flare.}
    \label{fig:4}
\end{figure*}

\begin{figure*}
    \centering
    \includegraphics[width=1\linewidth]{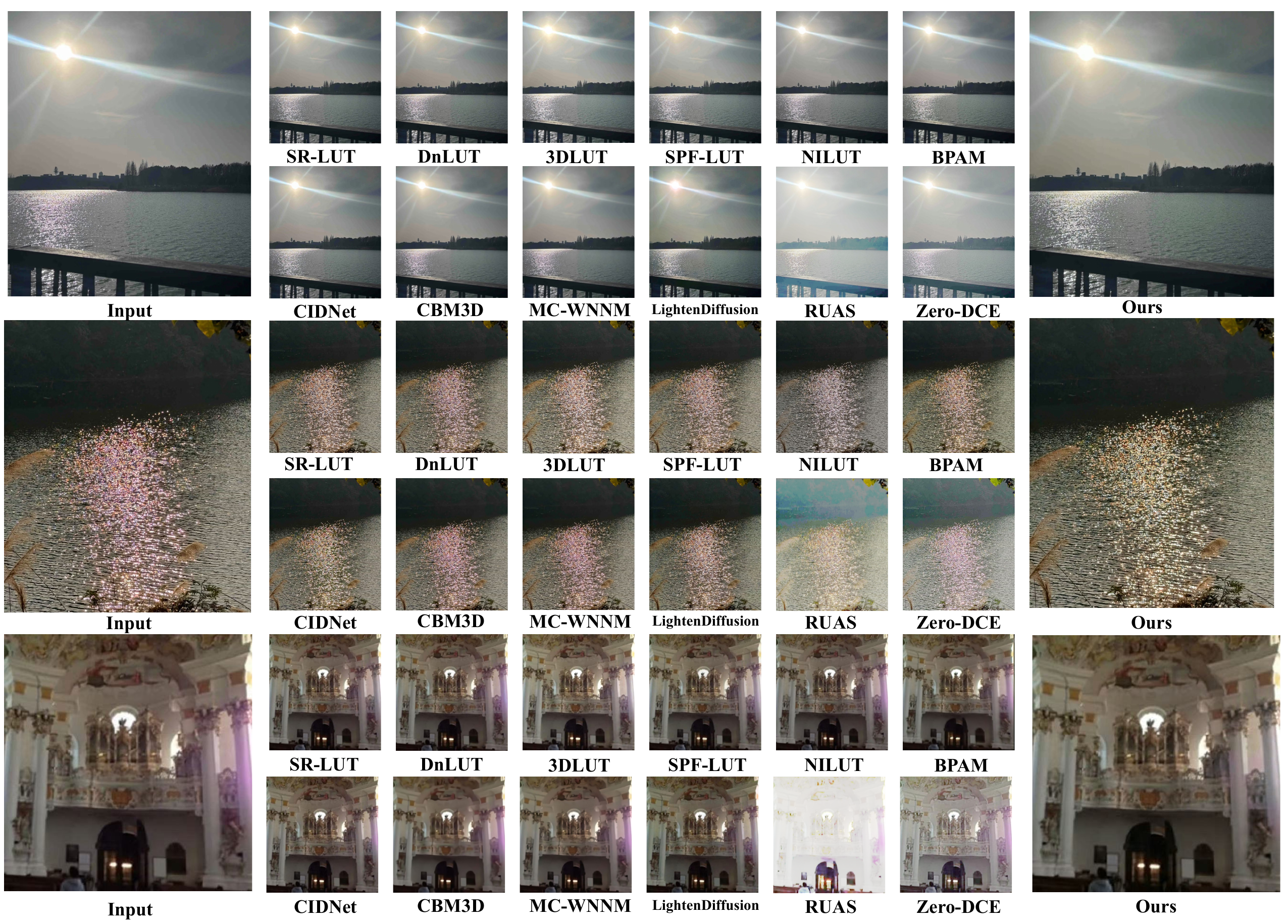}
    \caption{Visual comparison of all methods on a real-world photograph with purple flare.}
    \label{fig:5}
\end{figure*}

\begin{figure*}
    \centering
    \includegraphics[width=1\linewidth]{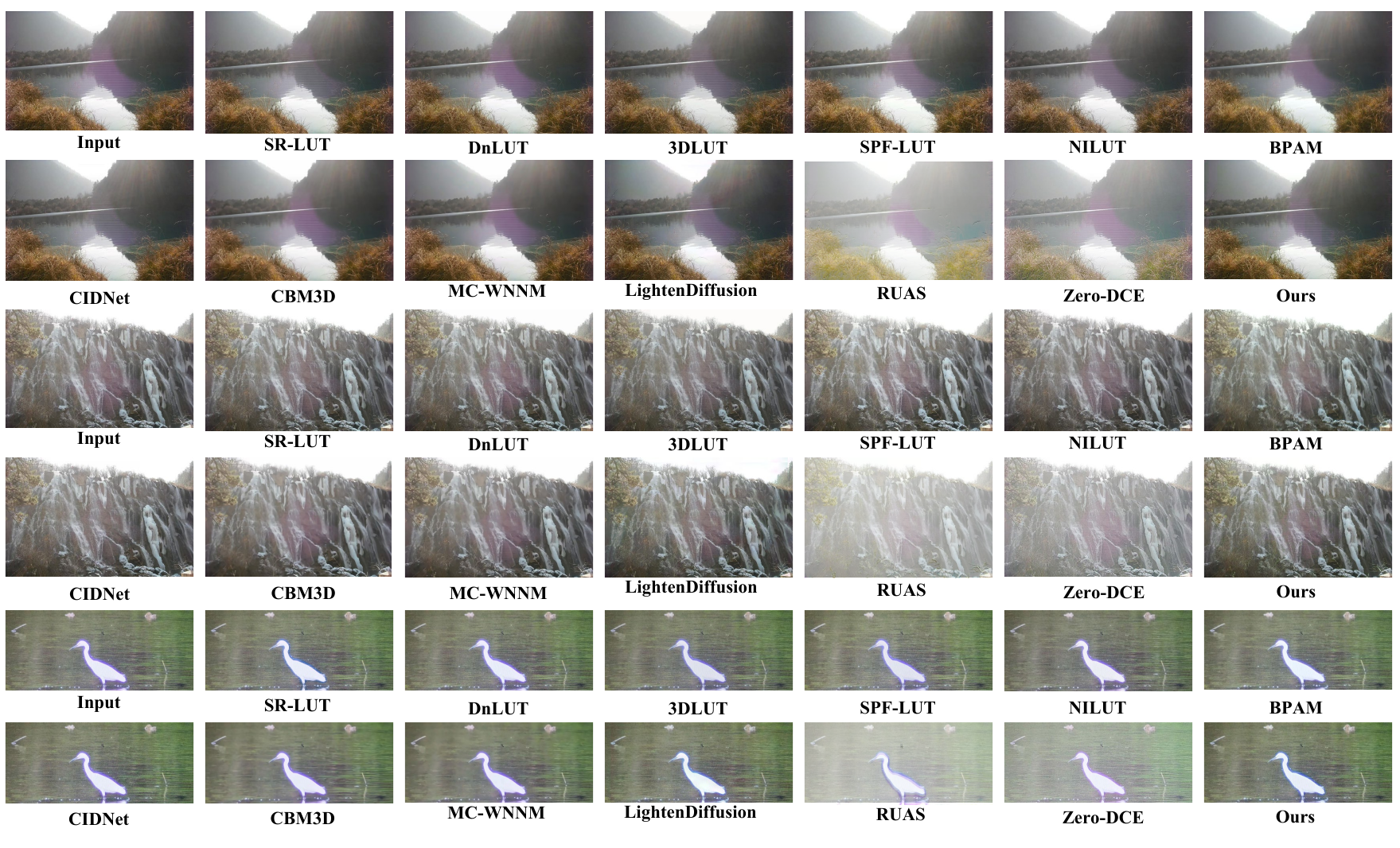}
    \caption{Visual comparison of all methods on a real-world photograph with purple flare.}
    \label{fig:6}
\end{figure*}

To verify the practical applicability and generalization ability of the model, we tested it on real-world images not included in the PFSD dataset. As shown in Figs. \ref{fig:4}, \ref{fig:5}, and \ref{fig:6}, the comparisons demonstrate that the proposed model effectively generalizes to real-world artifacts different from synthetic data, successfully removing purple flares while preserving the authenticity of the original photos. However, we note that in some rare and extreme cases, such as when the flare’s spectrum significantly deviates from typical purple flares (atypical colors, see the second and third examples in Fig. \ref{fig:6}) or when the flare is particularly large (see the first example in Fig. \ref{fig:4}), our model may fail to achieve perfect removal.

\end{document}